\documentclass[10pt,journal,compsoc]{IEEEtran}
\ifCLASSOPTIONcompsoc
  \usepackage[nocompress]{cite}
\else
  \usepackage{cite}
\fi

\ifCLASSINFOpdf
  \usepackage[pdftex]{graphicx}
\else
  \usepackage[dvips]{graphicx}
\fi
\usepackage{graphicx}
\usepackage{amsmath}
\usepackage{amssymb}
\usepackage{booktabs}
\usepackage{stfloats}
\usepackage{color}
\usepackage{times}
\usepackage[pagebackref, breaklinks, colorlinks, allcolors=black]{hyperref}
\usepackage{overpic}
\usepackage{bm}
\usepackage{tabu}
\usepackage{bbding}
\usepackage{multicol}
\usepackage{multirow}
\usepackage{scalerel,stackengine}
\usepackage[table]{xcolor}
\usepackage[accsupp]{axessibility}
\newcommand{\PAR}[1]{\vskip3pt \noindent{\bf #1~}}
\newcommand{\etal}{\textit{et al}.}
\usepackage{array}
\usepackage[capitalize]{cleveref}
\crefname{section}{Sec.}{Secs.}
\Crefname{section}{Section}{Sections}
\Crefname{table}{Table}{Tables}
\crefname{table}{Tab.}{Tabs.}
\usepackage{algorithm}
\usepackage{algorithmic}
 \usepackage[caption=false,font=footnotesize]{subfig}
\usepackage{url}

\newcommand{\greencheck}{{\Checkmark}}
\newcommand{\redx}{\XSolidBrush}

\begin{document}
\title{HiSC4D: Human-centered interaction and 4D Scene Capture in Large-scale Space Using Wearable IMUs and LiDAR}

\author{Yudi~Dai,
        Zhiyong~Wang, 
        Xiping~Lin,
        Chenglu~Wen,~\IEEEmembership{Senior~Member,~IEEE}, 
        Lan~Xu,
        Siqi~Shen,~\IEEEmembership{Senior~Member,~IEEE},
        Yuexin~Ma, 
        and~Cheng~Wang,~\IEEEmembership{Senior~Member,~IEEE}
\IEEEcompsocitemizethanks{
  \IEEEcompsocthanksitem Yudi Dai, Zhiyong~Wang, Xiping~Lin, Chenglu~Wen, Siqi~Shen and Cheng Wang are with the Fujian Key Laboratory of Sensing and Computing for Smart Cities and the School of Informatics, Xiamen University, Xiamen, FJ 361005, China. (E-mail: \{yudidai, wangzy, linxiping\}@stu.xmu.edu.cn; \{clwen, siqishen, cwang\}@xmu.edu.cn) 
  \IEEEcompsocthanksitem Lan Xu and Yuexin Ma are with the School of Information Science and Technology, ShanghaiTech University, Shanghai, 201210, China. (E-mail: \{xulan1, mayuexin\}@shanghaitech.edu.cn)}
\thanks{This work was supported in part by the National Natural Science Foundation of China under Grant 62171393, and Fundamental Research Funds for the Central Universities (No.20720220064). (Corresponding author: Chenglu Wen)}
}

\IEEEtitleabstractindextext{%
\begin{abstract}
We introduce HiSC4D, a novel \textbf{H}uman-centered \textbf{i}nteraction and \textbf{4D} \textbf{S}cene \textbf{C}apture method, aimed at accurately and efficiently creating a dynamic digital world, containing large-scale indoor-outdoor scenes, diverse human motions, rich human-human interactions, and human-environment interactions. 
By utilizing body-mounted IMUs and a head-mounted LiDAR, HiSC4D can capture egocentric human motions in unconstrained space without the need for external devices and pre-built maps. This affords great flexibility and accessibility for human-centered interaction and 4D scene capturing in various environments.
Taking into account that IMUs can capture human spatially unrestricted poses but are prone to drifting for long-period using, and while LiDAR is stable for global localization but rough for local positions and orientations, HiSC4D employs a joint optimization method, harmonizing all sensors and utilizing environment cues, yielding promising results for long-term capture in large scenes. 
To promote research of egocentric human interaction in large scenes and facilitate downstream tasks, we also present a dataset, containing 8 sequences in 4 large scenes (200 to 5,000 $m^2$), providing 36k frames of accurate 4D human motions with SMPL annotations and dynamic scenes, 31k frames of cropped human point clouds, and scene mesh of the environment.
A variety of scenarios, such as the basketball gym and commercial street, alongside challenging human motions, such as daily greeting, one-on-one basketball playing, and tour guiding, demonstrate the effectiveness and the generalization ability of HiSC4D.
The dataset and code will be publicly {available} for research purposes.

\end{abstract}

}

\maketitle

\IEEEdisplaynontitleabstractindextext

\IEEEpeerreviewmaketitle

\IEEEraisesectionheading{\section{Introduction}\label{sec:introduction}}
\IEEEPARstart{T}{he} overwhelming development of digital society greatly enriched people's lives from entertainment to education, and from commerce to art.
In many downstream areas, like autonomous driving, robotics, Augmented/Virtual/Mixed Reality, etc., there is a growing prominence of high-quality human-centric 4D content, which is essential for human action recognition, social-behavioral analysis,  scene perception, and AI-generated content area.
In the relevant research areas, it is a trend and cutting-edge to capture interacting 4D human motions in large real-world environments from an egocentric view. 
However, capturing such content accurately and robustly remains a substantial challenge, due to the complexities of capturing systems and challenging interaction patterns, including both human-to-human and human-environment interactions.

\begin{figure*}[!tb]
    \centering
    \includegraphics[width=0.995\linewidth]{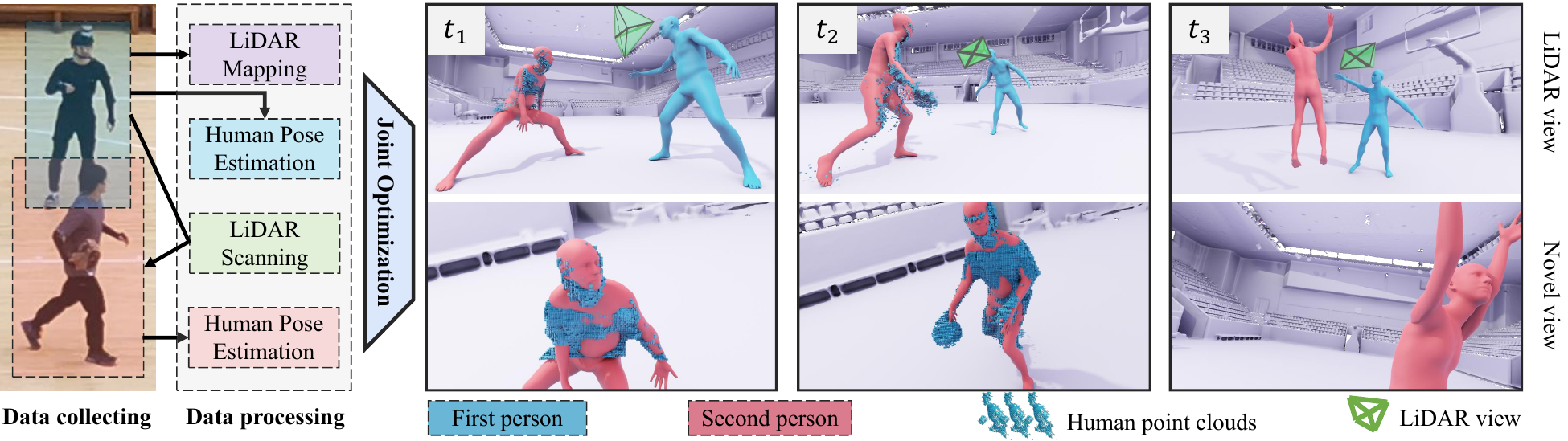}
    \vspace{-2mm}
    \caption{HiSC4D enables the capture of interacting human motions and large-scale scenes with a person equipped with a head-mounted LiDAR and both the first and the second person wearing IMUs. The LiDAR not only scans the surrounding environment but also captures detailed 3D annotations of the second person, ensuring accurate spatial information.}
    \label{fig:teaser}
    \vspace{-4mm}
 \end{figure*}

Inertial Measurement Unit (IMU) \cite{TimovonMarcard2017SparseIP} sensors have been frequently used in full-body human capture. It is scene-free and easy to mount on different human body parts, such as limbs, torso, and head. IMU can be of high sensitivity and short-term accurate motions. However, due to the sensor noise and offsets, the IMU-based motion capture system suffers from severe integration drift as the acquisition time increases. Therefore, building a drift-free motion-capturing system applied in large capture areas is the main challenge for IMU sensors. Some methods~\cite{Marcard_2018_ECCV, kaichi2020resolving, xu2017flycap, TrumbleBMVC17, dou2016fusion4d, hassan2019prox} utilize extra external RGB or RGBD cameras to improve the accuracy but result in limited capture space, human activities, and interactions. 
{PROX~\cite{hassan2019prox} leverages scene information to optimize human poses estimated from an RGBD camera. However, the static camera setup is not suitable for large-scale environments.}
HPS~\cite{guzov2021human} uses a head-mounted camera to complement IMUs in global localization. However, it requires pre-built maps and a huge image database for self-localization. 
EgoLocate~\cite{EgoLocate2023} removes the reliance on pre-built maps by utilizing the head-mounted camera for global localization and sparse scene reconstruction.
However, RGB-related methods tend to degenerate when human movement is highly dynamic or the scene lacks texture.

LiDAR has become widely used in the field of robotics and autonomous vehicles~\cite{zhang2014loam, caesar2020nuscenes, romero2017inlida, geiger2013vision, maddern20171}. Owing to its precise 3D geometric perception capability and remarkable performance in localization and mapping, LiDAR is attracting increasing attention for human motion estimation and capture~\cite{kim2019pedx, li2022lidarcap, Dai_2022_CVPR, ren2023lidar, Dai_2023_sloper4d, yan2023cimi4d}.
PedX~\cite{kim2019pedx} provides 3D poses of pedestrians by using SMPL~\cite{smpl2015loper} from third-person-view images in street scenes. 
LiDARCap~\cite{li2022lidarcap} utilizes a fixed LiDAR to capture the long-range 3D human pose but only in limited scenes. LIP~\cite{ren2023lidar} combines multiple complimentary IMUs with LiDAR to capture human motions in large-scale scenarios.
However, these approaches lack egocentric views and scene reconstruction.

Several methods~\cite{Xu2019Mo2Cap2RM, MiaoLiu20204DHB, EvonneNg2020You2MeIB, guzov2021human, SiweiZhang2021EgoBodyHB, Dai_2022_CVPR, EgoLocate2023} have explored the human motion capture from an egocentric view.
EgoBody~\cite{SiweiZhang2021EgoBodyHB} collects RGBD images and uses multiple pre-calibrated external RGBD sensors to provide 3D human ground truth, which requires tedious setup and the activities are limited in a single room.
HSC4D~\cite{Dai_2022_CVPR} firstly captures human-centric 4D scenes in large-scale indoor/outdoor spaces using a hip-mounted LiDAR with IMUs, but it only focuses on
the first-person pose estimation and can not capture human-human interactions in the scene.
SLOPER4D~\cite{Dai_2023_sloper4d} uses a head-mounted LiDAR with a camera to capture the second person's global human motion in large environments, yet lacks first-person motion capturing and human-human interaction.

To overcome these challenges, we introduce a novel approach termed \textbf{H}uman-centered \textbf{i}nteraction and \textbf{4D} \textbf{S}cene \textbf{C}apture (\textbf{HiSC4D}). This approach introduces a novel joint optimization method that integrates data from multiple sources, including IMU-based motion capture data from two individuals, LiDAR-based localization, human point clouds, and scene capturing. The objective of HiSC4D is the accurate creation of dynamic digital environments, faithfully capturing human motions and social interactions in diverse settings, encompassing both indoor and outdoor scenarios.
Relying solely on body-mounted sensors, HiSC4D is capable of capturing human interactions in expansive environments without spatial or pose limitations. By cooperating LiDAR sensor, our approach not only mitigates IMU drifting but also remains insensitive to lighting conditions. Additionally, our LiDAR-based localization and 3D scene reconstruction eliminates the reliance on pre-built maps, streamlining the setup process and ensuring adaptability for a wide range of real-world scenes.
Finally, utilizing the multi-modal data including point cloud, IMUs motions, and SLAM trajectories, we introduce a multi-stage joint optimization framework, which incorporates the designed scene-aware physical constraints,  to obtain naturally interacting human motions within realistic scenes.

To promote the research on egocentric human interaction within large-scale scenes and to facilitate downstream applications, we introduce the HiSC4D dataset — a multi-modal egocentric human motion dataset. This dataset consists of eight sequences captured across four diverse and expansive scenes, spanning from 200 to 5,000 square meters. It provides around 36k frames of global 4D human motion data and dynamic scenes, 31k frames of cropped human point clouds from an egocentric view, comprehensive 3D ground truth, and dense scene mesh.
As depicted in \cref{tab:data_desc}, the HiSC4D dataset encompasses a range of scenarios, including a basketball gym, a multi-story building, a commercial street, and an outdoor area on campus. These scenarios involve a variety of challenging human interaction activities, such as ball toss, basketball training, stair climbing, and guided tours.
Through comprehensive experimentation, our approach has been thoroughly evaluated. The accurate depiction of human poses and natural interactions in large scenes demonstrates the efficacy and generalization capability of our dataset. Additionally, the inclusion of first-person view data introduces novel challenges that provide valuable insights for the development of more robust approaches.
Our contributions are summarized as follows:
\begin{itemize}
    \item We propose a novel method that enables the capture of Human-centered interaction and 4D scenes in expansive spaces. This approach leverages cues between head-worn LiDAR and body-mounted IMUs, allowing us to reconstruct human motions with social interactions in a dynamic digital environment, significantly broadening the scope of spatial, motion, and interaction capture.

    \item We introduce a multi-stage joint optimization framework, which integrates the processed multi-modal data including human point clouds, IMU motions, and LiDAR SLAM results, by scene constraints. This optimization results in natural global human motions within large scenes.
    
    \item We present a novel egocentric human motion dataset in large scenes. This dataset includes daily and challenging motions with social interactions, as well as diverse scenarios. It provides various types of multi-modal data, such as global human motions with SMPL parameter annotations, 4D scene sequences, cropped human point clouds, and scenes with mesh format for research purposes.

\end{itemize}

\section{Related work}
\label{sec:Related_work}
This section presents an overview of research works related to our proposed HiSC4D system. We roughly divide the related methods into LiDAR/IMU-based 3D Human Pose Estimation, Human Localization and Scene Mapping, and Human-centered Social Interaction capture.

\PAR{LiDAR/IMU-based 3D Human Pose Estimation.} 
IMU sensors have been widely used to capture human motions~\cite{roetenberg2007moven, vlasic2007practical, SIP, DIP:SIGGRAPHAsia:2018, yi2022physical}. 
However, IMU-based methods suffer from severe drift over time. To improve the pose estimation accuracy, some methods~\cite{Marcard2016HumanPE, dou2016fusion4d, Malleson3DV17, huang2017towards, Marcard_2018_ECCV, kaichi2020resolving} utilize extra external RGB or RGBD cameras as a remedy. 
Helten \etal~\cite{Helten:2013} combined two RGBD cameras with IMUs to perform local pose optimization.
HybridFusion~\cite{zherong2018} has achieved more accurate motion-tracking performance by combining an RGBD camera with multiple IMUs. 
{Total capture~\cite{TrumbleBMVC17} fuses multi-view video with IMU data to accurately estimate 3D human pose in a lab environment. However, it has a limited sensing range and cannot be used outdoors.}
3DPW~\cite{Marcard_2018_ECCV} uses a single hand-held RGB camera and IMUs to optimize human pose for a certain period of frames simultaneously.
Constraints from external cameras assist in recovering more accurate 3D poses but result in limited capture space, human activities, and interactions. 
HPS~\cite{guzov2021human} uses a first-view head-mounted camera to self-localize the 3D pose from IMUs to the scene. However, it requires pre-built maps and an image database for self-localization. 
{PIP~\cite{yi2022physical} combines physics-based motion optimization with sparse inertial motion capture, using only six IMUs to capture human motion in real-time. However, it lacks interaction with real-world environments.}
HSC4D~\cite{Dai_2022_CVPR} achieved promising results for long-term human motion and scene capture by using only body-mounted IMUs and a hip-mounted LiDAR. {However, it cannot capture other humans in the scene. 
Recently, EgoLocate~\cite{EgoLocate2023} estimates accurate human pose and localization using six IMUs and a head-mounted camera, it can also reconstruct sparse scenes without needing pre-scanning. 
}
Our method is complementary to HSC4D in that {we capture accurate 3D human poses of the second person, who interacts with the first person}.  

\PAR{Human Localization and Scene Mapping.}
Human self-localization aims at estimating the 6-DoF of the human subject with carrying devices. The received signal strength (RSS) fingerprinting-based methodologies~\cite{abbas2019wideep, lemic2014infrastructure, alarifi2016ultra} are widely used for indoor human localization. However, these methods need external receivers and are limited to the indoor space. Some image-based methods~\cite{kendall2015posenet, radwan2018vlocnet++, wang2020atloc} regress locations directly from a single image with a pre-built map. Still, the scene-specific property makes it hard to generalize to unseen scenes. Some methods integrate IMU as an aid sensor \cite{shan2020lio, oleynikova2015real} to improve accuracy. 
LiDAR is widely used in simultaneous localization and mapping \cite{zhang2014loam} \cite{shan2018lego} \cite{bosse2012zebedee} \cite{lin2020loam}  due to its robustness and low drift. To localize the human subject, LiDAR are designed as backpacked \cite{Liu2010IndoorLA, 8736839, Karam2019DesignCA} and hand-held\cite{bauwens2016forest}. 
LiDAR has been successfully applied in indoor localization~\cite{wang2018single, peng2017lidar} and outdoor localization~\cite{yin2018locnet, uy2018pointnetvlad, yu2021deep, li2019net} scenes. 
However, LiDAR-based localization and mapping systems are big pieces of equipment, that affect human motion and not being suitable for daily activities. Our capturing system is lightweight and uses a head-mounted LiDAR, achieving self-localization and mapping in large indoor and outdoor scenes. 
LOAM~\cite{zhang2014loam} is a real-time odometry and mapping method greatly boosting LiDAR-based SLAM research. 
Some methods~\cite{shan2018lego, wang2021, lin2020loam, jiao2021robust} further improve LOAM mapping for specific scenes and sensors. 
LeGO-LOAM\cite{shan2018lego} is a ground-optimized version, which requires keeping the LiDAR horizontal. 
LiDAR-based methods tend to fail when the $Z$-axis jitters severely. 
To address the drift problem and improve robustness, more sensors, such as visual sensors\cite{zhang2015visual, shin2020dvl, seo2019tight}, IMU\cite{shan2020lio, geneva2018lips, opromolla2016lidar}, or both\cite{deilamsalehy2016sensor, zuo2019lic, shan2021lvi}, have been integrated in mapping task. 
Among LiDAR-based methods and datasets, a lot of attention has been paid to autonomous driving \cite{kitti} \cite{gskim-2020-mulran} or robotics from the third-person view, and they usually do not focus on humans. With the aid of our designed capturing system, we captured an HiSC4D dataset containing LiDAR human motions from the first-person's egocentric view, as well as the 3D pose ground truth. 

\PAR{Human-centered Social Interaction capture}
In recent years, there has been an increasing interest in first-person vision research~\cite{Xu2019Mo2Cap2RM, MiaoLiu20204DHB, EvonneNg2020You2MeIB,guzov2021human, SiweiZhang2021EgoBodyHB, Dai_2022_CVPR}, in which some focus on social interaction predicting~\cite{AlirczaFathi2012SocialIA, MichaelSRyoo2013FirstPersonAR}, and some focus on first-person 3D joints/pose estimation \cite{HaoJiang2016SeeingIP, TakaakiShiratori2011MotionCF}.
However, most of them usually lack 3D pose ground truth or cannot reconstruct the 3D environment. 
The You2Me~\cite{EvonneNg2020You2MeIB} captures the interaction between the first person (wearing a chest-mounted camera) and the second person, and infers the first person's poses from the interaction. However, it lacks 3D scene context.
4DCapture~\cite{MiaoLiu20204DHB} collects egocentric videos and reconstructs second-person 3D human body meshes, but does not have first-person body data and reliable 3D annotations.
HPS~\cite{guzov2021human} reconstructs the human body pose using IMUs and a head-mounted camera in large 3D scenes but has few social interactions and heavily relies on the pre-built map. 
EgoBody~\cite{SiweiZhang2021EgoBodyHB} pre-scans the scene with an iPhone and uses multiple depth sensors to provide 3D human motion ground-truth, but has tedious multiple sensors setup and calibration and its social interactions are limited in a small room. 
HSC4D~\cite{Dai_2022_CVPR} achieves large human-scene modeling without scene-prior and contributes a novel LiDAR - IMUs setting for first-person vision data capturing. However, LiDAR on the back is unable to record social interactions with the second person.
By upgrading HSC4D and solving the mapping problem with a built-in IMU of the LiDAR, our HiSC4D successfully reconstructed interactions with the second person and provided accurate SMPL annotations. We believe HiSC4D data will foster further research in the egocentric vision in large scenes, especially the unique LiDAR modality.

\section{Methodology}
\label{sec:method}

\begin{figure*}[!htb]
    \centering
     \includegraphics[width=0.95\linewidth]{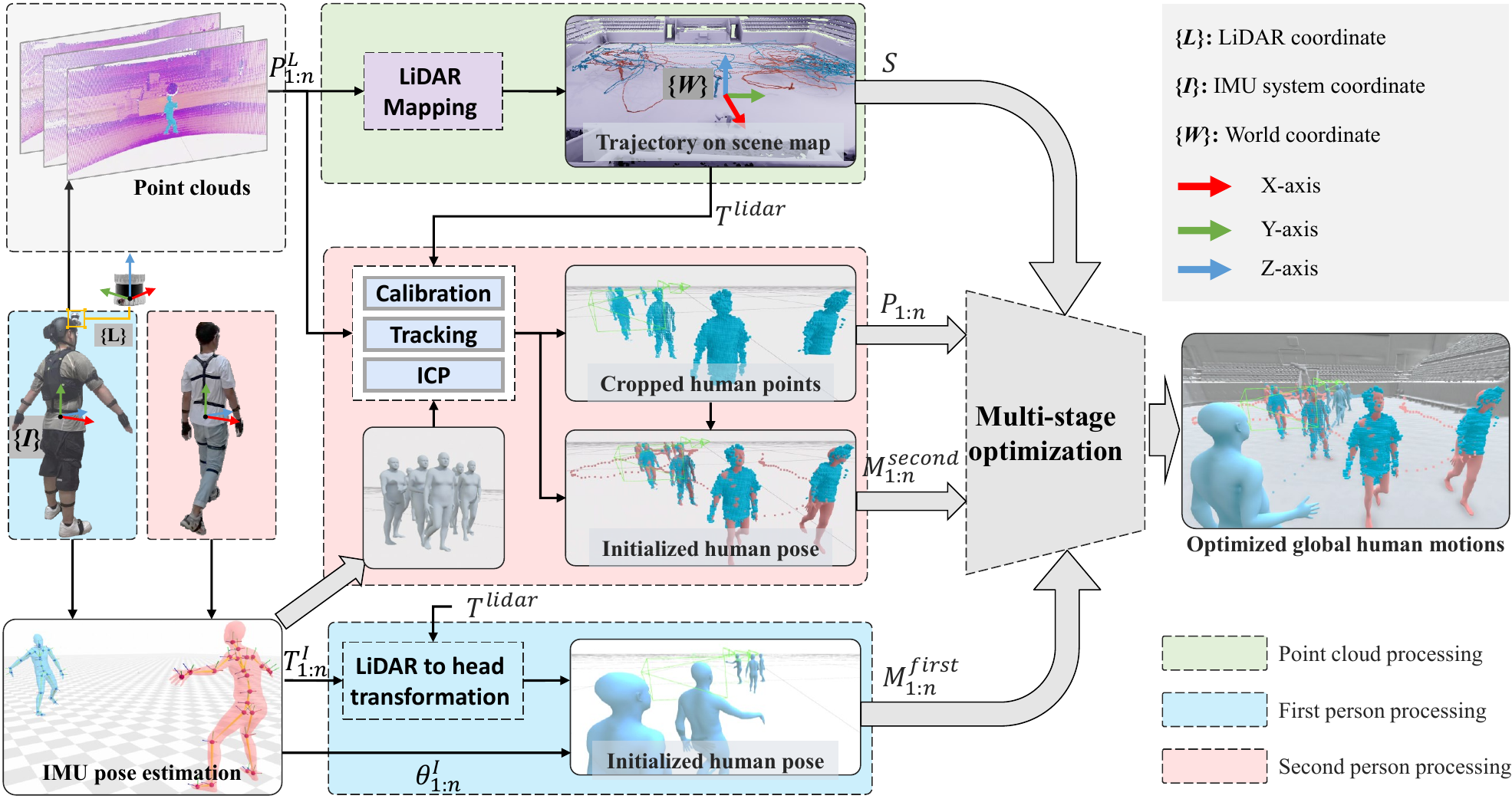}
     \vspace{-2mm}
  
     \caption{\textbf{The pipeline of HiSC4D.}
     The HiSC4D pipeline consists of LiDAR mapping, first-person motion processing, second-person motion processing, and a multi-stage joint optimization process. This comprehensive pipeline enables the capture and reconstruction of the dynamics of two humans and the scene, resulting in accurate localization and natural human interactions in large-scale environments.
     }
     \label{fig:method}
    \vspace{-4mm}
\end{figure*}

To address these challenges, we introduce our innovative approach —— Human-centered Interaction and 4D Scene Capture (HiSC4D). As shown in \cref{fig:method}, after data has been collected, it begins with the data {processing} stage (\cref{subsec:processing}). Here, we process the raw IMU data and LiDAR point cloud data to obtain essential components: IMU-based human motion, and LiDAR-based localization and mapping. Subsequently, by utilizing LiDAR localization and segmented human point cloud, we initialize the human motion's global positions. Finally, as detailed in \cref{sec:optimization}, our approach employs a multi-stage optimization strategy to iteratively get accurate human-centered interaction and scene capture.

\PAR{Notations.} We use the right subscript $k$, where $k\in~Z^+$, to indicate the index of a frame, and the right superscript, $I$ or $L$ or $W$ (defaulting to $W$), to indicate the coordinate system to which the data belongs.
The $k$-th frame point cloud is represented by $P_k^L$, and the 3D scene is represented as \bm{$S$}. 
Here, We use the Skinned Multi-Person Linear (SMPL)\cite{smpl2015loper} body model to map the $k$-th frame pose parameters $T_k$, $\theta_k$, and $\beta$ to its motion representation $M_k$. This mapping function is defined as $M_k = \varPhi (T_k, R_k, \theta_k, \beta)$, where $M_k\in\mathbb{R}^{6890\times3}$ represents the triangle vertices. The global translation $T_k \in \mathbb{R}^3$ represents the translation of the SMPL model, while the pose parameter is composed of the pelvis joint's orientation $R_k \in \mathbb{R}^{1\times3}$ relative to the start frame and the rotations of the other 23 joints  $\theta_k~\in \mathbb{R}^{23\times3}$ relative to their parent. The constant parameter $\beta \in \mathbb{R}^{10}$ represents the human body shape, which is obtained by using the IPNet\cite{bhatnagar2020ipnet} to optimize the human mesh captured by an iPhone14Pro.

\PAR{Coordinates system definition.} 
{We define three coordinate systems used in our paper: 1) IMU coordinate system \{$I$\} follows the left-hand rule. The origin is located at the pelvis joint of the first SMPL model with the $X/Y/Z$ axis pointed to the right/upward/forward relative to the human body. 2) LiDAR coordinate system \{$L$\} follows the left-hand rule. The origin is at the bottom center of the sensor with the $X/Y/Z$ axis pointed to the forward/left/left/top of the LiDAR. 3) Global/World coordinate system \{$W$\} follows the right-hand rule for consistency with \{$L$\}. It's the scene's coordinate we manually define.}

\subsection{Data processing}
\label{subsec:processing}
The {processing} is to acquire the initialized human motions and LiDAR data for optimization. First, we estimate the LiDAR ego-motion and create a 3D scene map using the point clouds. Next, we acquire the 3D human motion output using our inertial MoCap system. Finally, we relocalize the human motions using LiDAR localization and segmented human point clouds.

\PAR{LiDAR-Inertial Localization and Mapping.}
Building a global consistency map and estimating accurate trajectory is difficult for the LiDAR-only method in this scene because the LiDAR jitters as the human walks and jumps. Incorporating an IMU will solve this problem by compensating for the motion distortion in a LiDAR scan and providing a good initial pose. Our LiDAR-Inertial SLAM method contains two modules, one is iterated Kalman filter-based LiDAR-Interial odometry\cite{Xu2022FASTLIO2FD}, and another is factor-graph-based loop closure optimization\cite{gtsam}\cite{Kim2018ScanCE}. By employing this LiDAR-Interial SLAM method, we estimate the ego-motion of LiDAR and build the global consistency 3D scene map $\bm{S}$ with $P_k^L, k\in~Z^+$ in \{$L$\}. 
We first use the iterated Kalman filter-based state estimation module to estimate the full LiDAR state by registering raw points in a scan to the map points, then using a mapping module incrementally add the new points from each scan to an ikd-Tree\cite{Cai2021ikdTreeAI}.
Second, we use scan context\cite{Kim2018ScanCE} as a place recognition method to find the loop closure position. Then, we use factor graph\cite{gtsam} to perform loop closure optimization to generate a global consistency map.
Finally, LiDAR's ego-motion $T^W$ and $R^W$, and the global consistency scene map $S$ are computed. The mapping function is denoted as:
\begin{equation}
    {T^{lidar}}_{1:n}, {R^{lidar}}_{1:n}, \bm{S} = \mathcal{F} (P_{1:n}^L, R_{WL}).
    \label{equa:mapping}
\end{equation}

\PAR{IMUs Pose Estimation.}
To obtain each person's initial $n$ frame motions $M_{1:n}$ in world coordinate, we transform the $R^I$ and $T^I$ from $\{I\}$ to $\{W\}$. The calculation is defined as:
\begin{equation}
    \begin{split}
        M_{1:n} = \varPhi (R_{WI}T^I_{1:n}, R_{WI} R^I_{1:n}, \theta_{1:n}, \beta),
    \end{split}
\label{equa:imu_pose}
\end{equation}
\noindent where $T^I$, $R^I$, and $\theta^I$ are provided by the commercial MoCap product. The global translation of the initial human motion $M_{1:n}$ is very inaccurate due to the IMU's severe drifting for long-period capture. Additionally, the IMUs' pose estimation algorithm assumes that the person is walking on a flat plane without height changes. In order to address this issue, we will utilize the LiDAR localization result $T^{lidar}$ for the first person's global translation calculation and the cropped LiDAR human point clouds for the second person's global translation calculation.

\label{subsec:initialization}
\PAR{The First-person Global Localization.}
We set the pelvis localization $T_{pelvis}$ to represent the subject's translation. The pelvis's position can be calculated directly from the head's joint position, which can be inferred from the LiDAR's localization. The calculation is defined as follows:
\begin{equation}
	\begin{split}
    T_{pelvis} & = T_{ph} + T_{hl} + T_{lidar},\\
    \end{split}
\label{equa:troot}
\end{equation}
where $T_{lidar}$ is the head-mounted LiDAR translation, $T_{hl}$ is the LiDAR to head offset, and $T_{ph}$ is the head to root joint offset. The $T_{ph}$ is calculated as: 
\begin{equation}
    T_{ph} = J_{pelvis}(\theta, \mathcal{\beta}) - J_{head}(\theta, \mathcal{\beta}),
\end{equation}
where $J(\theta, \mathcal{\beta})$ maps from pose $\theta$ and shape $\beta$ to the body joints. Since the LiDAR is rigidly attached to the head, their rotation offsets are the same, $T_{hl}$ is defined as: 
\begin{equation}
	\begin{split}
    T_{hl} & = R_H(\theta) t_{hl},
    \end{split}
\label{equa:lidarhead}
\end{equation}
where $R_H$ maps from pose $\theta$ to head rotation, and $t_{hl}$ is a constant LiDAR-to-head offset with zero rotation, which can be estimated at the first frame, $t_{hl} = (R_{H}(\theta_0))^\top (T_{hl})_0$.
To get the head's global orientation $R_H(\theta)$, the kinematic chain of SMPL is iteratively traversed through its parent joint until the pelvis joint,
\begin{equation}
R_{H} (\theta)= {\displaystyle \prod_{i \in \mathcal{P}_\mathrm{Head}}} {\theta^{(i)}}^\lor,
\label{eq:head}
\end{equation} 
where $\mathcal{P}_\mathrm{Head}$ is an ordered list of all the parents to the head joint,  $^\lor$ operator converts axis-angle representations to rotation matrices.

By solving the orthogonal Procrustes problem~\cite{gower2004procrustes} to find the matrix $ \Omega,\ \mathrm{subject}\ \mathrm{to}\ \Omega ^{T}\Omega =I$, which most closely maps $T^{W}$ to $T_{pelvis}$, we further refine the calibration matrix $R_{WI}$ with the rotation matrix's yaw. Specifically,
\begin{equation}
    \begin{split}
        \varDelta R &=\arg \min _{\Omega }\|\Omega T^{W}-T_{pelvis}\|_{F}, \\
        R_{WL}'\,\! &= \varDelta R_{yaw} R_{WL},
    \end{split}
\label{equa:comput_similarity}
\end{equation} 
where $\|\cdot \|_{F}$ denotes the Frobenius norm and $\varDelta R$ is the solution for the optimal value of ${\Omega }$  that minimizes the norm squared ${||\Omega T^{W}-T_{pelvis}\|_{F}^{2}}$.
With the refined $R_{WL}'\,\!$, we update the $M^W$ as per~\cref{equa:imu_pose} and then update the new $T_{pelvis}$ as per~\cref{equa:troot}. At last, we get the first person's global localization $T_{pelvis}$. So far, the first-person data $M$ has been initialized.

\PAR{The Second-person Global Localization.}
To extract human body points from LiDAR frames for the second person, we use PV-RCNN~\cite{shi2020pv} and PC3T~\cite{wu20213d} to track the human body points. PV-RCNN detects all human 3D bounding boxes in LiDAR frames. However, due to the domain gap and the significant perspective difference between the autonomous driving dataset and our dataset, these state-of-the-art detection and tracking techniques result in many false positives and missing. 
To address this, we propose a method to fuse the tracking result given by PC3T and the IMUs translation, filtering the incorrect 3D boxes and completing the missing parts, as well as optimizing the IMUs trajectory. 
Specifically, for all candidate boxes in every frame, we choose the closest one, $B_{ij}$, to the updated IMU translation, where $B_{ij}$ denotes the $j\text{-th}$ box in $i\text{-th}$ frame. Then we crop the $P_i$'s points in $B_{ij}$ as human body points $\mathcal{P}_i$ and calculate the center of $B_{ij}$ as the new human translation.
Then we use Hidden Points Removal (HPR) \cite{katz2007direct} to remove the invisible mesh vertices based on the LiDAR's perspective. Finally, we use Iterative closest point (ICP)~\cite{segal2009generalized} to register the visible vertices to $\mathcal{P}_i$.
Following the steps in~\cref{equa:comput_similarity}, we calculate the $\varDelta R$ that maps the $T^{W}$ to $T'\,\!$ and then refine the $R_{WI}$ for the second person. With the refined $R_{WL}'\,\!$, we update the IMUs motions.

\subsection{Joint optimization}
\label{sec:optimization}
\begin{figure}[!htb]
    \centering
    \includegraphics[width=0.99\linewidth]{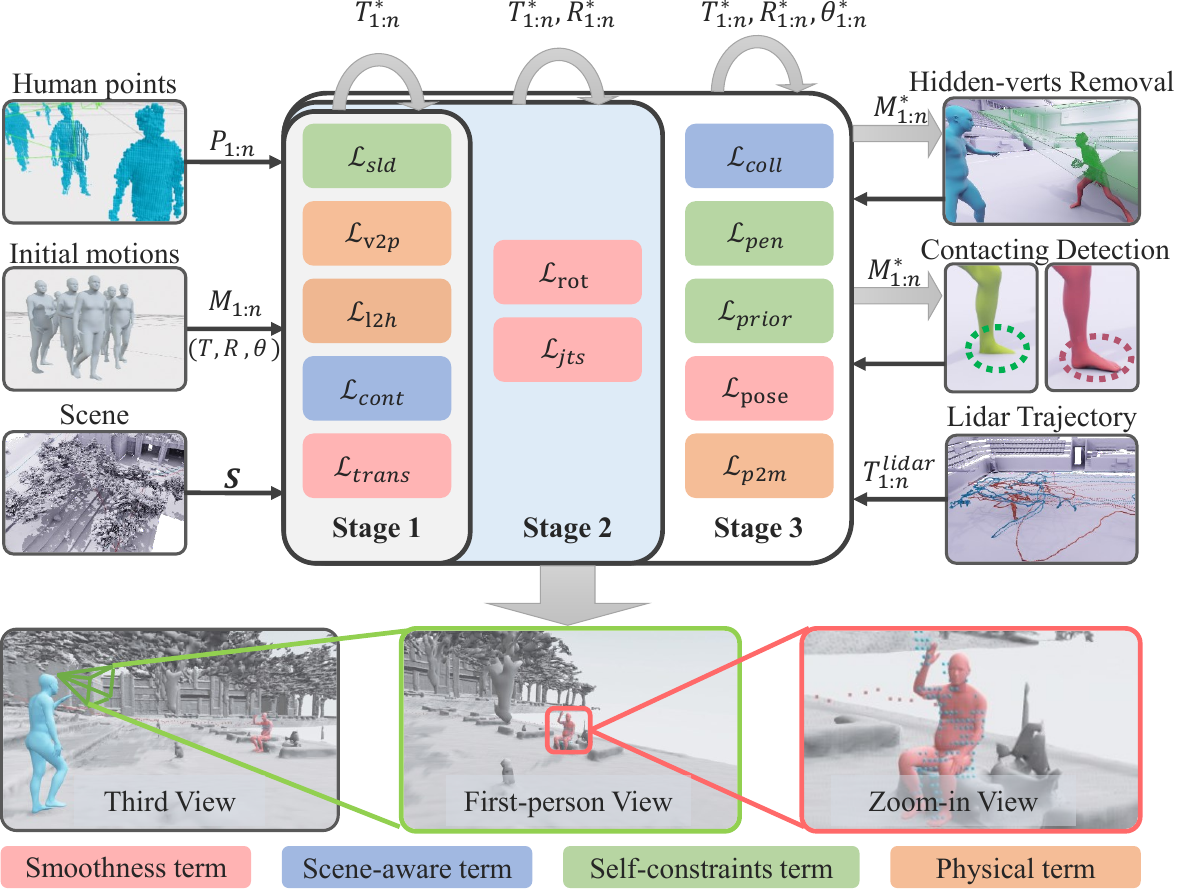}  
    \vspace{-2mm}
    \caption{The mult-stage optimization pipeline for dual-person motions. 
    \textbf{Stage 1}: optimizing the global translation $T$ only.
    \textbf{Stage 2}: optimizing $T$ and the global rotation $R$.
    \textbf{Stage 3}: optimizing $T$, $R$, and $\theta$. 
    The pipeline takes $P_{1:n}$, initial human motions $M_{1:n}$, and the scene $S$ as input, outputs accurate and scene-plausible motions in large environments.}
    \vspace{-4mm}
    \label{fig:data_chart}
 \end{figure}

In this subsection, we tackle the problem of optimizing human motion for each person given two sequences of 3D human motion, denoted as $\{{\bm{M}_{1:n}^I}\}^{first}$ and $\{{\bm{M}_{1:n}^I}\}^{second}$ in the inertial coordinate $I$. 
Additionally, we utilize the scene mesh $\bm{S}$ in the world coordinate $W$, and a sequence of human point clouds denoted as ${P_{1:n}^L}$ in the LiDAR coordinate $L$. 
The objective is to output the optimized human motion, ${\bm{M}_{1:n}^*}$, in the world coordinate $W$ for each person.
The optimization can be formulated as follows:
\begin{equation}
	\begin{split}
    \bm{M_{1:n}^*} = &\arg \min _{\bm{M}}\mathcal{L} (M_{1:n}, \bm{S}, P_{1:n}),
    \\    
    \mathcal{L} = & \mathcal{L}_{smt} + \mathcal{L}_{self} + \mathcal{L}_{scene} + \mathcal{L}_{phy}.
    \end{split}
\end{equation}
\PAR{Optimization Strategy.} 
To achieve accurate and realistic human motion ${\bm{M}_{1:n}^*}$ and ensure faster convergence, we employ a multi-stage optimization strategy, as shown in \cref{fig:data_chart}. In the first stage, we optimize the global translation parameter exclusively. Subsequently, in the second stage, both global translation and rotation are optimized. Finally, in the last stage, all SMPL parameters are optimized simultaneously. We incorporate specific constraints into the optimization process, which are categorized as follows: smoothness terms ($\mathcal{L}_{smt}$), human self-constraints terms ($\mathcal{L}_{self}$), scene-aware terms ($\mathcal{L}_{scene}$), and physical terms ($\mathcal{L}_{phy}$). These constraints are derived from the human point clouds and the LiDAR trajectory.
\subsubsection{The smoothness terms} 
To ensure smoothness in both temporal and spatial aspects of the movements. The $\mathcal{L}_{smt}$ term incorporates the following constraints:
(1) the translation term $\mathcal{L}_{trans}$, promoting smoothness in the global translation of the human body by minimizing the pelvis's acceleration.
(2) The rotation term $\mathcal{L}_{rot}$, encourages smooth transitions in the global rotation of the human body by minimizing the pelvis's angular acceleration.
(3) The body pose smoothness $\mathcal{L}_{pose}$, ensuring smoothness in the body pose, maintaining consistency throughout the motion sequence by minimizing each pelvis-relative joint's angular velocity.
(4) The body joints terms $\mathcal{L}_{jts}$, promoting smoothness in the movements of each pelvis-relative joint by minimizing its acceleration.
As a result, the $\mathcal{L}_{smt}$ is expressed as:
\begin{equation}
	\begin{split}
        \mathcal{L}_{smt} = 
        \lambda_{trans} \mathcal{L}_{trans} + 
        \lambda_{rot} \mathcal{L}_{rot}
        + \lambda_{pose} \mathcal{L}_{pose} + 
        \lambda_{jts} \mathcal{L}_{jts},
    \end{split}
\end{equation}
where $\lambda_{trans}$, $\lambda_{rot}$, $\lambda_{pose}$, and $\lambda_{jts}$ are coefficients of corresponding loss terms. These terms are detailed as follows:
\begin{equation}
	\begin{split}
         & \mathcal{L}_{trans} = 
        \frac{1}{k-2}\sum_{i=1}^{k-2}\|T_{i+2} - 2T_{i+1} + T_{i}\|_2^2,\\ 
        & \mathcal{L}_{\text {rot}} =  
        \frac{1}{k-1}\sum_{i=1}^{k-1} 
        \|R_{i+1} - R_{i}\|_{2}^2, \\
        & \mathcal{L}_{\text {pose}} =  
        \frac{1}{k-1}\sum_{i=1}^{k-1} 
        \|\theta_{i+1} - \theta_{i}\|_{2}^2, \\
        & \mathcal{L}_{jts} =  
        \frac{1}{k-2} \sum_{i=1}^{k-2}\|{J(M_{i+2}^*)} - 2J(M_{i+1}^*) + J(M_{i}^*)\|_2^2,\\ 
    \end{split}
\end{equation}
where 23 pelvis-relative joints are regressed from the motions by $J(M^*_i) \in \mathbb{R}^{23 \times 3}$.

\subsubsection{Self-constraints terms}
\label{subsec:selfcons}

The $\mathcal{L}_{\text{self}}$ term combines the constraints of foot sliding constraint ($\mathcal{L}_{sld}$), pose prior constraint ($\mathcal{L}_{prior}$), and self-penetration constraint ($\mathcal{L}_{pen}$) to improve the local pose quality. It ensures realistic foot positions, aligns the body pose with initial estimates, and prevents mesh interpenetration. The $\mathcal{L}_{self}$ is expressed as:
\begin{equation}
	\begin{split}
        \mathcal{L}_{self} = 
        \lambda_{sld} \mathcal{L}_{sld} +
        \lambda_{prior} \mathcal{L}_{prior} +
        \lambda_{pen} \mathcal{L}_{pen},
    \end{split}
\end{equation}
where $\lambda_{sld}$, $\lambda_{prior}$, and $\lambda_{pen}$ are coefficients of these loss terms.

\PAR{Foot Sliding Constraint.} The $\mathcal{L}_{sld}$ reduces the foot sliding between consecutive stable foot vertices, resulting in more natural and smooth human movements. The $\mathcal{L}_{\text{sld}}$ is defined as the Euclidean distance between the stable feet in successive frames:
\begin{equation}
	\begin{split}
        \mathcal{L}_{\text{sld}}=\frac{1}{k-1}
        \sum_{i=2}^{k}  
        \|\mathbb{E}(\mathcal{F}_{j})-\mathbb{E}(\mathcal{F}_{j-1})\|_{2},
    \end{split}
    \label{equ:sld}
\end{equation}
\noindent 
where $\mathbb{E}$ represents the mean of a given set of values.

\PAR{Pose Prior Constraint.} 
To ensure optimization begins with a reliable initial value, we introduce the pose prior constraint $\mathcal{L}_{prior}$, which enforces the $\theta$ to closely align with the initial estimation from the IMUs, as introduced in~\cref{subsec:processing}. 
However, the IMU poses are pseudo values and may deviate from the true values due to drifting. To address this issue, the pose prior constraint gradually decreases as the number of iterations $iter$ increases. This constraint is defined as follows:
\begin{equation}
	\begin{split}
        \mathcal{L}_{\text {prior}} =
        \frac{1}{iter}\frac{1}{k}
        \sum_{i=1}^{k} 
        \|\theta_{i}^* - \theta_{i}^I\|_{2}^2.
    \end{split}
\end{equation}

\PAR{Self-penetrating Constraint.} 
To address self-penetrations in human motion $M$, we introduce the self-penetrating constraint $\mathcal{L}_{\text{pen}}$. 
To efficiently identify self-penetrating vertex pairs, we adopt a strategy that partitions the body mesh $M$ into separate regions encompassing the torso, arms, hands, legs, and head. Subsequently, we search vertex pairs from distinct body regions, thereby substantially decreasing computational cost while retaining the capacity to detect potential self-penetrations. The self-penetrating constraint $\mathcal{L}_{\text{pen}}$ can be formulated as:

\begin{equation}
	\begin{split}
    \mathcal{L}_{{pen}} = & \frac{1}{k}
    \sum_{i=1}^{k}
    \sum_{A\in {M_i}} 
    \sum_{\substack{B\in {M_i}}}
    \sum_{\mathbf{a} \in A} \max\left(0, \left(\mathbf{a} - \mathbf{b}\right) \cdot \mathbf{n}_b\right), \\
    & \quad s.t.~\mathbf{b} = \arg\min_{\mathbf{b} \in B} \|\mathbf{a} - \mathbf{b}\|.
	\end{split}
\end{equation}
\noindent
where $A$ and $B$ represent different body regions, and vertices $\mathbf{a}$ and $\mathbf{b}$ belong to their respective regions. The $\mathbf{n}_b$ denotes the normalized normal vector at $\mathbf{b}$.
The self-penetration is determined by the positive dot product between the vector $(\mathbf{a} - \mathbf{b})$ and $\mathbf{n}_b$.

\subsubsection{Scene-aware terms}
By incorporating environmental cues, the $\mathcal{L}_{scene}$, composed as foot contact loss ($\mathcal{L}_{cont}$) and human-scene collision constraint ($\mathcal{L}_{coll}$), plays a crucial role in enhancing the contextual authenticity of the human motions, resulting in realistic interaction with their surroundings. The $\mathcal{L}_{scene}$ is formulated as follows:
\begin{equation}
    \begin{split}
        \mathcal{L}_{scene} = 
        \lambda_{cont} \mathcal{L}_{cont} +
        \lambda_{coll} \mathcal{L}_{coll}.
    \end{split}
\end{equation}

\PAR{Foot Contact Constraint.} 
The foot contact loss, denoted as $\mathcal{L}_{cont}$, is defined as the distance from a stable foot to the nearest ground surface. To determine the foot state, we compare the movements of the left and right foot for each consecutive foot vertex in ${M}_i$. If the movement of a foot is less than 2$cm$ and less than the movement of the other foot, the foot is marked as stable.
The stable foot vertices in $M_i$ are denoted as $\mathcal{F}_{i}$. The $\mathcal{L}_{cont}$ is formulated as:
\begin{equation}
	\begin{split}
        \mathcal{L}_{\text {cont}}=\frac{1}{k}
        \sum_{j=1}^{k}
        \sum_{v \in \mathcal{F}_j}
        \frac{1}{|\mathcal{F}_j|} \min_{\hat{p} \in S}\|v - \hat{p}\|_{2}^{2}.
    \end{split}
    \label{equ:cont}
\end{equation}
\noindent

\PAR{Human-scene Collision Constraint.} 
By punishing the vertices in SMPL mesh penetrating the scene mesh, the $\mathcal{L}_{coll}$ ensures that the human mesh remains collision-free with the 3D scene during optimization. 
Similar to the method of detecting penetrations in \cref{subsec:selfcons}, we find the closest point $S_v$ on scene $S$ for every vertex $v$ in $M$, if the dot product between the distance vector from $\mathbf{v}$ to $S_v$ and the normal vector at $S_v$ is positive, it is identified as a term to be optimized. The calculation is formulated as:
\begin{equation}
	\begin{split}
    \mathcal{L}_{{coll}} = & \frac{1}{k}
    \sum_{i=1}^{k}
    \sum_{\mathbf{v} \in M_i} 
    max\left(0, \left(S_v - \mathbf{v}\right) \cdot \mathbf{n}_{S_v}\right), 
	\end{split}
\end{equation}
\noindent
where $S_v$ is the projection of $v$ on $S$ that follows the minimum distance direction and $\mathbf{n}_{S_v}$ is the normalized normal vector of $S_v$.

\subsubsection{Physical terms.}
To leverage the accurate localization provided by the LiDAR sensor and the rich depth and positional information from the point cloud, we introduce a comprehensive physical loss term $\mathcal{L}_{\text{physical}}$. This loss term combines the constraints imposed by the LiDAR-based head localization, the viewpoint-based vertices-to-point loss $\mathcal{L}_{\text{v2p}}$, and the point-to-mesh loss $\mathcal{L}_{\text{p2m}}$.
The $\mathcal{L}_{phy}$ is formulated as:
\begin{equation}
	\begin{split}
        \mathcal{L}_{phy} = 
        \lambda_{l2h} \mathcal{L}_{l2h} +
        \lambda_{v2p} \mathcal{L}_{v2p} +
        \lambda_{p2m} \mathcal{L}_{p2m},
    \end{split}
\end{equation}
where $\lambda_{l2h}$, $\lambda_{v2p}$, and $\lambda_{p2m}$ are coefficients of these loss terms.

\PAR{LiDAR-head Constraint.} 
The LiDAR sensor is rigidly attached to the head of the human body. {Consequently}, the Euclidean distance between the optimized position of the head and the LiDAR translation $T_{\text{lidar}}$ should be constrained to a constant value. This constraint ensures that the head remains in a fixed position relative to the LiDAR.
\begin{equation}
	\begin{split}
        \mathcal{L}_{l2h}=\frac{1}{k}\sum_{i=1}^{k} 
        \|(T_{lidar})_{i} + {(T_{hl})}_{i} - {J}_{head}{(V_{i})}\|_{2},
    \end{split}
    \label{equ:head}
\end{equation}
where $T_{hl}$ is the vector from the LiDAR to SMPL head joints, which is defined in~\cref{equa:lidarhead}. 
Note that this constraint is only used for the LiDAR wearer (the first person).

\begin{figure}[!tb]
    \centering
    \includegraphics[width=0.99\linewidth]{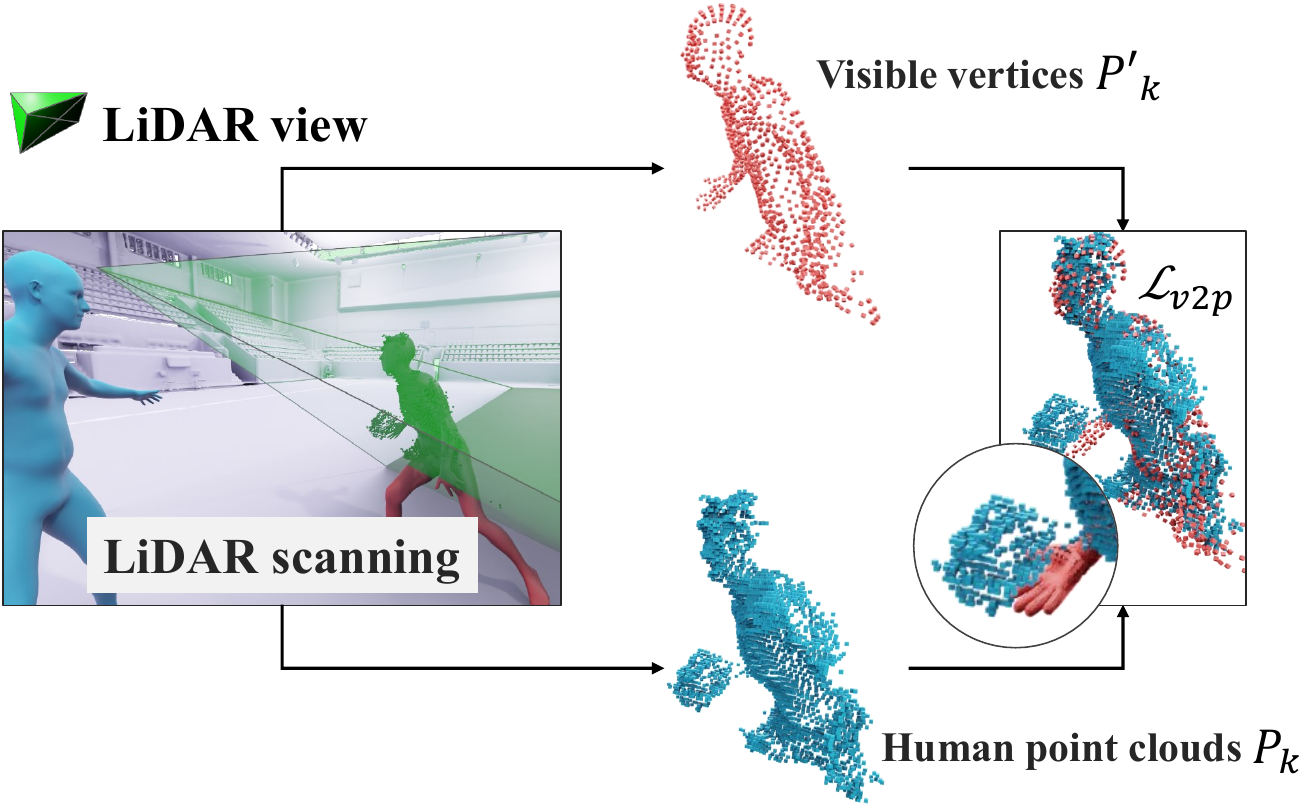}
    \vspace{-2mm}
    \caption{\textbf{The vertices to point constraint pipeline.} First, we remove the second person's (red) invisible vertices from the LiDAR view, then resample the visible SMPL vertices to align the LiDAR's resolution. Finally, we apply the $\mathcal{L}_{v2p}$ to the second person.}
     \label{fig:verts_removel}
    \vspace{-3mm}

\end{figure}

\PAR{Vertices to Point Constraint.} 
The point cloud $P_{1:k}$ from the moving LiDAR provides strong prior depth information. As a result, fitting the SMPL mesh to the human point cloud would produce satisfactory results.
However, traditional registration methods like ICP (Iterative Closest Point) may not be ideal {for aligning} sparse and partial human points with dense and full SMPL meshes. These methods typically rely on a large overlapping area and similar point set densities to achieve accurate results.
To address this issue, we propose a viewpoint-based vertices to point loss term $\mathcal{L}_{v2p}$, which specifically tackles the challenges associated with this type of data.
As shown in \cref{fig:verts_removel}, we first remove the hidden SMPL vertices from the LiDAR's viewpoint, ensuring the resampled SMPL mesh's density matches the human point cloud's density.
And then we sample points, denoted as $P'\,\!_{1:k}$, from the remaining faces by LiDAR resolution.
Finally, we iteratively minimize the Chamfer Distance (CD) between the point cloud $P'\,\!_{1:k}$ and the sampled SMPL mesh points $P_{1:k}$.
It is important to note that we only consider the distance from $P'\,\!_{1:k}$ to $P_{1:k}$.
This is because $P_{1:k}$ typically has more points than $P'\,\!_{1:k}$. For example, in \cref{fig:verts_removel}, $P_{k}$ only includes the basketball, while $P'\,\!_{k}$ only includes the right hand. If we were to consider the CD from $P_{k}$ to $P'\,\!_{k}$ as well, it could potentially lead to pushing the hand to intersect with the basketball, which is undesirable.
The constraint is regularized with the following equation: 
\begin{equation}
	\begin{split}
        \mathcal{L}_{v2p} = 
        \frac{1}{k}\sum_{i=1}^{k} 
        \sum_{\hat{p'\,\!}\in P'\,\!_{i}} \frac{1}{|P'\,\!_{i}|} \min_{\hat{p} \in P_{i}}\|\hat{p} - \hat{p'\,\!}\|_{2}^{2}.
    \end{split}
\end{equation}

\PAR{Point to Mesh Constraint.} 
In the later stages of optimization, when the results become more stable, we introduce the point-to-mesh loss term $\mathcal{L}{p2m}$ to further improve the non-rigid alignment between the point cloud and the SMPL mesh. 
However, to avoid including distant non-body points such as the ball or clothes, we introduce a threshold value for the distance. The point-to-mesh loss $\mathcal{L}{p2m}$ is then defined as the sum of distances between each point in the point cloud and its closest point on the mesh surface, only considering distances below the threshold.
Let $P_{1:k}$ be the point cloud and $M_{1:k}$ be the optimized SMPL mesh. The point-to-mesh loss $\mathcal{L}_{p2m}$ can be expressed as:

\begin{equation}
	\begin{split}
        \mathcal{L}_{p2m} = 
        \frac{1}{k}\sum_{i=1}^{k} 
        \sum_{\hat{p}\in P_{i}} \frac{1}{|P_{i}|} \min ( \min_{\hat{v} \in M_{i}}\|\hat{p} - \hat{v}\|_{2}^{2}, ~ 0.15).
    \end{split}
\end{equation}

\section{Datasets}
\label{sec:Overview}
The HiSC4D system collects human-centric 4D data focused on human behavior, including both first and second-person motions, as well as scene data. 
In this section, we introduce {data collection, data processing, dataset statistics, and dataset comparison.}

\begin{table*}[!tbp]
	\centering
    \renewcommand\arraystretch{1.2}
    \caption{\textbf{Statistical Analysis of the Dataset.} 'F-Traj.' represents the length of the first-person trajectory, 'S-Traj.' represents the length of the second-person trajectory, and 'Visible' indicates the number of human point frames visible in each sequence. The area size is measured in square meters.}
	\centering
    \vspace{-2mm}

    \begin{tabular}{@{}l|l|lllll|l@{}}
        \toprule
        \textbf{Num.} & \textbf{Sequence} & \textbf{F-Traj.} & \textbf{S-Traj.} & \textbf{Area size} & \textbf{Frames} & \textbf{Visible} & \textbf{Description} \\ \midrule
        \rowcolor[HTML]{D9D9D9} 
        S1 & building\_001 & 182m & 200m & 2,360 & 3925 & 3077 & Walking together from inside to the outside of a building. \\
         &  &  &  &  &  &  & Attempting to access a locked   classroom on the second floor \\
        \multirow{-2}{*}{S2} & \multirow{-2}{*}{building\_002} & \multirow{-2}{*}{136m} & \multirow{-2}{*}{239m} & \multirow{-2}{*}{380} & \multirow{-2}{*}{4519} & \multirow{-2}{*}{3807} & and go   down together. \\
        \rowcolor[HTML]{D9D9D9} 
        \cellcolor[HTML]{D9D9D9} & \cellcolor[HTML]{D9D9D9} & \cellcolor[HTML]{D9D9D9} & \cellcolor[HTML]{D9D9D9} & \cellcolor[HTML]{D9D9D9} & \cellcolor[HTML]{D9D9D9} & \cellcolor[HTML]{D9D9D9} & Two-person interactions involving ball  tossing, chair moving, \\
        \rowcolor[HTML]{D9D9D9} 
        \multirow{-2}{*}{\cellcolor[HTML]{D9D9D9}S3} & \multirow{-2}{*}{\cellcolor[HTML]{D9D9D9}building\_003} & \multirow{-2}{*}{\cellcolor[HTML]{D9D9D9}97m} & \multirow{-2}{*}{\cellcolor[HTML]{D9D9D9}135m} & \multirow{-2}{*}{\cellcolor[HTML]{D9D9D9}180} & \multirow{-2}{*}{\cellcolor[HTML]{D9D9D9}3251} & \multirow{-2}{*}{\cellcolor[HTML]{D9D9D9}3017} & and greetings in a building lobby. \\
        S4 & basketball\_001 & 255m & 386m & 420 & 7966 & 5999 & Two-person basketball training   sessions in an indoor gym, \\
        \cellcolor[HTML]{D9D9D9}S5 & \cellcolor[HTML]{D9D9D9}basketball\_002 & \cellcolor[HTML]{D9D9D9}96 & \cellcolor[HTML]{D9D9D9}242 & \cellcolor[HTML]{D9D9D9}420 & \cellcolor[HTML]{D9D9D9}3517 & \cellcolor[HTML]{D9D9D9}3216 & including passing, dribbling, shooting,   defense, and \\
        S6 & basketball\_003 & 131m & 75m & 420 & 2931 & 2494 & one-on-one play. \\
        \rowcolor[HTML]{D9D9D9} 
        S7 & campus\_001 & 204m & 300m & 2,400 & 5487 & 5254 & Two-person touring during a campus walk. \\
        S8 & street\_001 & 369m & 415m & 5,600 & 4391 & 4122 & Walking together along a pedestrian shopping street at night. \\ \bottomrule
        \end{tabular}
     \vspace{-3mm}

	\label{tab:data_desc}
\end{table*}

\subsection{Data Collection}

\begin{figure}[!tb]
    \centering
     \includegraphics[width=0.98\linewidth]{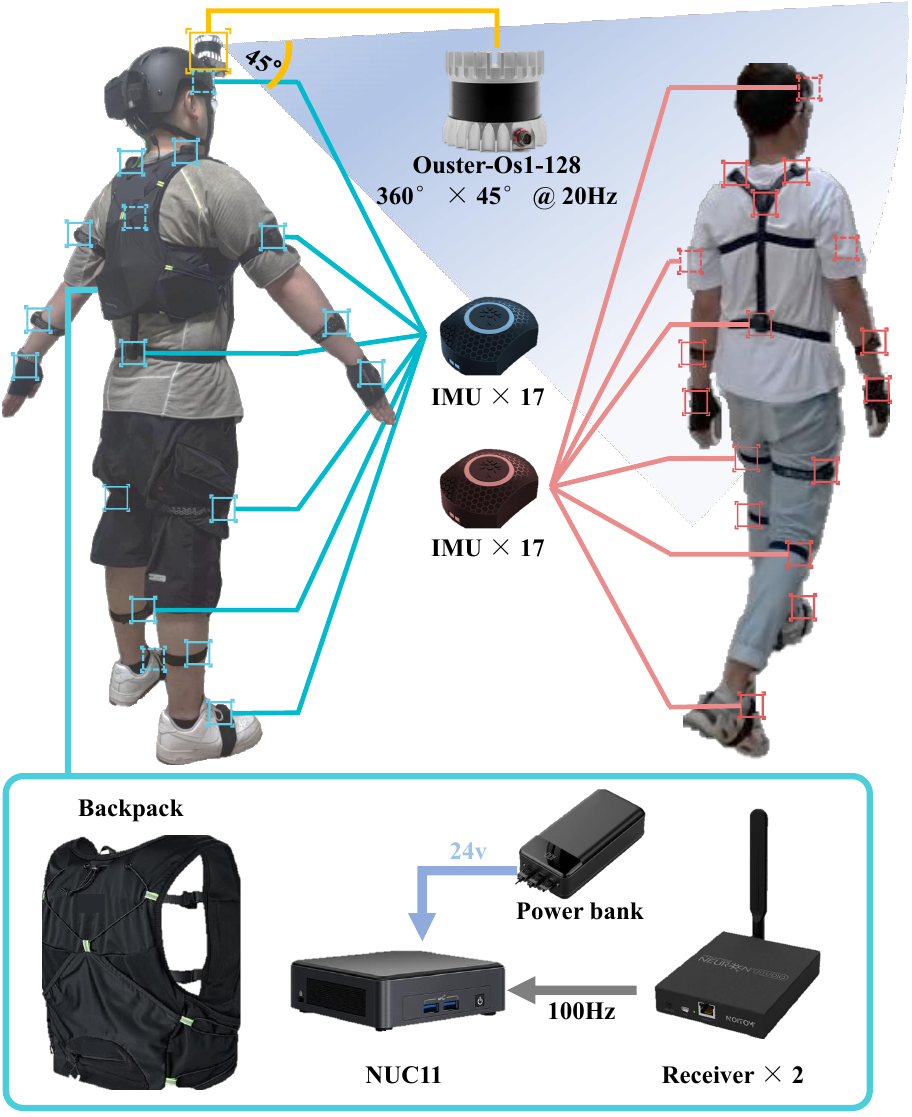}
     \vspace{-2mm}
     \caption{Design of the capturing system: Both individuals are equipped with 17 body-attached IMU sensors. The first person also wears a head-mounted LiDAR sensor and a backpack, which houses receivers for both sets of IMUs, a mini-computer, and a mobile power bank.}
     \label{fig:system_design}
     \vspace{-4mm}

\end{figure}
\PAR{Hardware.}
We use a 128-beam Ouster-os1 LiDAR to acquire 3D point clouds and Noitom's inertial MoCap product PN Studio to obtain human motion. Each suit of PN Studio uses 17 IMUs attached to the body limbs and two receivers to acquire data. Receivers and the LiDAR are connected to an Intel NUC11 mini-computer and use 24V mobile power to charge the LiDAR and the computer.
The LiDAR is rigidly connected to the helmet via a bracket; the data is transmitted to a mini-computer (NUC11) through the cable behind the helmet. We assume that LiDAR and the head have a rigid transformation. The LiDAR has a 360° horizon field of view (FOV) and a 45° vertical FOV. To make the people entering the LiDAR's FOV as much as possible, we tilt down the LiDAR around 45°. 
To ensure a lightweight and easy-to-use capturing system, we modified all cables and delicately designed the hardware position, making the battery ($200mm\times120mm\times50mmm$), two receivers ($130mm\times100mm\times30mmm$), and the mini-computer ($117mm\times112mm\times57mmm$) stored in a cross-country backpack. Finally, as \cref{fig:system_design} shows, we get a concise capturing system, requiring only one key to activate each sensor.

\PAR{Data Capture.} 
Before data capturing, the LiDAR wearer stands in A-pose facing or parallel to a large real-world plane marker with a flat face, such as a wall or a square column.  The second person stands in front of the LiDAR wearer, 5 meters away, facing the same direction as the LiDAR wearer. {Once the data capturing begins, both individuals will freely move around the environment and engage in interactions. When ready to end the capturing, both individuals will return to their starting positions.}

\subsection{Data Processing}

{After data collecting, we first calibrate the LiDAR and IMU data to the world coordinates, then synchronize all the data. This allows us to apply the proposed method described in \cref{sec:method} to the data.}

\PAR{Calibration}
The LiDAR wearer's right, front, and up directions are designated as the scene's $X$, $Y$, and $X$ axis, respectively. The midpoint of the subject's ankle projection on the horizontal plane is set as the origin. After data collection, we employ the RANSAC~\cite{schnabel2007efficient} to detect the planes in the first frame point cloud, manually selecting the ground plane, and obtaining its normal vector, detected as $g=[g_1,g_2,g_3]^\top$. Additionally, we fit a plane for the plane marker and compute its normal $m=[m_1, m_2, m_3]$, oriented in the LiDAR wearer's right-hand direction. 
The calculation of the coarse calibration matrix $R_{WL}$ from the first frame LiDAR point cloud to the world coordinate $\{W\}$ is as follows:
\begin{equation}
    R_{WL} = 
    \begin{bmatrix}
    s_1 & s_2 & s_3 & 0\\
    m_1 & m_2 & m_3 & 0.2\\
    g_1 & g_2 & g_3 & h \\
    0   & 0   & 0   & 1
    \end{bmatrix},
\end{equation} 
where $[s_1, s_2, s_3]^\top = m \times g$ and $h$ represents the LiDAR wearer's height. {Specifically, this coarse} $R_{WL}$ {is further refined in \cref{subsec:processing}, which is used in the optimization process}. Based on the definition of IMU coordinate system \{$I$\}, the coarse transformation matrix $R_{WI}$ that transforms it to the \{$W$\} is defined as: 
\begin{equation}
    R_{WI} = \begin{bmatrix}
    -1 & 0 & 0 & 0\\
    0  & 0 & 1 & 0\\
    0  & 1 & 0 & 0\\
    0 & 0 & 0 & 1
    \end{bmatrix}.
\end{equation}

\PAR{Synchronization.} 
LiDAR data is captured at a rate of 20 frames per second (FPS), while IMU poses of each human are captured at a rate of 100 FPS. 
At the beginning and end of the capture, we automatically detect peaks in IMU and LiDAR trajectories by having the LiDAR wearer perform a jump in place. 
However, due to the discrete nature of the trajectory, there is always a system error affecting the accuracy. To obtain a more accurate peak time $t_{peak}$, we employ a curve-fitting approach. This approach leverages the fact that, in the vicinity of the highest point during the jump, the trajectory heights $H_{peak}$ exhibit a quadratic relationship with time. Therefore, we utilize the location of the trajectory around the detected peaks and its timestamps for curve fitting."
\begin{equation}
	\begin{split}
        E(t_{peak}) & = \sum_{h}^{h \in H_{peak}} \|\frac{1}{2}~g~(t(h)-t_{peak})^2 - h\|_{2}, \\
        {t_{peak}}^* & =  arg \min_{t_{peak}} E(t_{peak}), 
    \end{split}
\end{equation}
\noindent where $g$ is the Gravity parameter, which is 9.8 $m/s^2$, and $t(h)$ is the time of $h$.
Finally, we synchronize the LiDAR and IMUs based on the corresponding peak timestamps and resample the IMU data to 20 FPS to match the LiDAR frame rate.

\PAR{Processing.} 
{We then processed the data using our proposed method, resulting in SMPL pose, shape, and global translations for both individuals, human point clouds for the second person, and 3D scene maps. 
}

\subsection{Dataset Statistics}
\label{sec:dataset}

{With the raw data and the processed results, we present a novel human-centered dataset called HiSC4D, which contains 4D interacting human motions and scenes. 
As shown in \cref{tab:data_desc}, HiSC4D comprises over 30k frames of eight sequences of multi-modal data captured in four real-world large indoor and outdoor environments: a building lobby, an indoor basketball gym, the campus area, and a pedestrian shopping street at night. The captured area spans from 200 to 5000 $m^2$, each showcasing a variety of long-term human motion and interactions, including daily activities, human social interactions, as well as sports.}
The interactions involve activities such as ball tossing, chair moving, greetings, basketball training sessions, campus touring, and shopping.

\PAR{Human Point Clouds Sequences.}  
To contribute to the LiDAR-based human pose estimation research and benchmark this task, HiSC4D further provides 25,000 frames of visible segmented human points for the second person, as well as our optimized 3D human body pose and shape annotations in SMPL template as ground truth. 
Due to the first person's head motion and the limitation of LiDAR's FOV, the second person is occluded sometimes. We split each sequence into multiple segments based on the consistency of visible frames.

\PAR{3D Scenes.} 
HiSC4D provides the dense 3D point cloud map reconstructed from the LiDAR as well as the colorful point cloud map from a Terrestrial Laser Scanner (Trimble TX5) for better visualization. These scenes include an indoor basketball stadium, a multi-story office building, a tree-lined campus area, and a pedestrian shopping street at night. The basketball gym is over 500 $m^2$. The indoor and outdoor area size of the multi-floor building is larger than 2,400 $m^2$. 
The pedestrian shopping street area is over 5600 $m^2$, and our capturing trajectory reaches 415 meters.

\begin{table*}[!htbp]
    \caption{ \textbf{Comparison with existing egocentric 3D human pose datasets.} "Global" refers to the human poses in global coordinates. "1P" denotes the first-person, "2P" denotes the second-person, and "\# Type" specifies whether the data is real or synthetic.
    }
     \centering

\vspace{-2mm}

\begin{tabular}{l|c|cc|cccc|cc|ccc}
    \toprule
    \multicolumn{1}{l|}{\multirow{2}[4]{*}{Dataset}} & \multicolumn{1}{c|}{\multirow{2}[4]{*}{\# Frame}} & \multicolumn{2}{c|}{Setting} & \multicolumn{4}{c|}{Pose} & \multicolumn{2}{c|}{Scene} & \multicolumn{3}{c}{Modality} \\
    \cmidrule(lr){3-4}\cmidrule(lr){5-8}\cmidrule(lr){9-10} \cmidrule(lr){11-13}     & \multicolumn{1}{c|}{} & \# Type & Outdoor & 1P   & 2P   & Global & Interaction & 3D scene & w/o pre-scanning & LiDAR & RGB  & IMU \\
    \midrule
    EgoCap~\cite{rhodin2016egocap} & 30k  & Real & \redx & \greencheck & \redx & \redx & \redx & \redx & N/A  & \redx & \greencheck & \redx \\
    Mo2Cap2~\cite{Xu2019Mo2Cap2RM} & 530k & Real & \greencheck & \greencheck & \redx & \redx & \greencheck & \redx & N/A  & \redx & \greencheck & \redx \\
    You2Me~\cite{EvonneNg2020You2MeIB} & 151k & Real & \redx & \greencheck & \greencheck & \redx & \greencheck & \redx & N/A  & \redx & \greencheck & \redx \\
    xR-EgoPose~\cite{tome2019xr} & 383  & Synthetic & \greencheck & \greencheck & \redx & \redx & \redx & \redx & N/A  & \redx & \greencheck & \redx \\
    HPS~\cite{guzov2021human} & 300k & Real & \greencheck & \greencheck & \redx & \greencheck & \redx & \greencheck & \redx & \redx & \greencheck & \greencheck \\
    EgoBody~\cite{SiweiZhang2021EgoBodyHB} & 153k & Real & \redx & \greencheck & \greencheck & \greencheck & \greencheck & \greencheck & \redx & \redx & \greencheck & \redx \\
    HSC4D~\cite{Dai_2022_CVPR} & 10k  & Real & \greencheck & \greencheck & \redx & \greencheck & \redx & \greencheck & \greencheck & \greencheck & \redx & \greencheck \\
    UnrealEgo~\cite{akada2022unrealego} & 459k & Synthetic & \greencheck & \greencheck & \redx & \redx & \redx & \redx & N/A  & \redx & \greencheck & \redx \\
    SLOPER4D~\cite{Dai_2023_sloper4d} & 100k & Real & \greencheck & \redx & \greencheck & \greencheck & \redx & \greencheck & \greencheck & \greencheck & \greencheck & \greencheck \\
    \midrule
    HiSC4D (Ours) & 36k  & Real & \greencheck & \greencheck & \greencheck & \greencheck & \greencheck & \greencheck & \greencheck & \greencheck & \redx & \greencheck \\
    \bottomrule
    \end{tabular}%
    
     \label{tab:data_compare}
    \vspace{-4mm}

 \end{table*}

\subsection{Dataset Comparison}
\label{subsec:dataset_compare}
Humans perceive and interact with the world from a first-person perspective. As shown in \cref{tab:data_compare}, many 3D human pose/motion datasets attempted to capture data from this viewpoint. However, they usually lack 3D pose ground truth or environmental context and are restricted to laboratory environments, with limitations including the absence of global settings or human interactions.
In contrast, HiSC4D, augmented with calibrated LiDAR point clouds and IMU data, addresses these limitations by enabling precise global human pose estimation and scene reconstruction. It is proficient in capturing diverse social interactions in real-world scenes, providing intricate 3D human pose ground truth.
HiSC4D's integration of interaction and global elements in real settings positions it as a versatile tool with the potential to explore various research dimensions, including social interaction analysis, long-term human motion analysis, and human activity recognition.

\section{experiments}
\label{sec:experiments}
\begin{figure*}[!htb]
   \centering
   \includegraphics[width=0.98\linewidth]{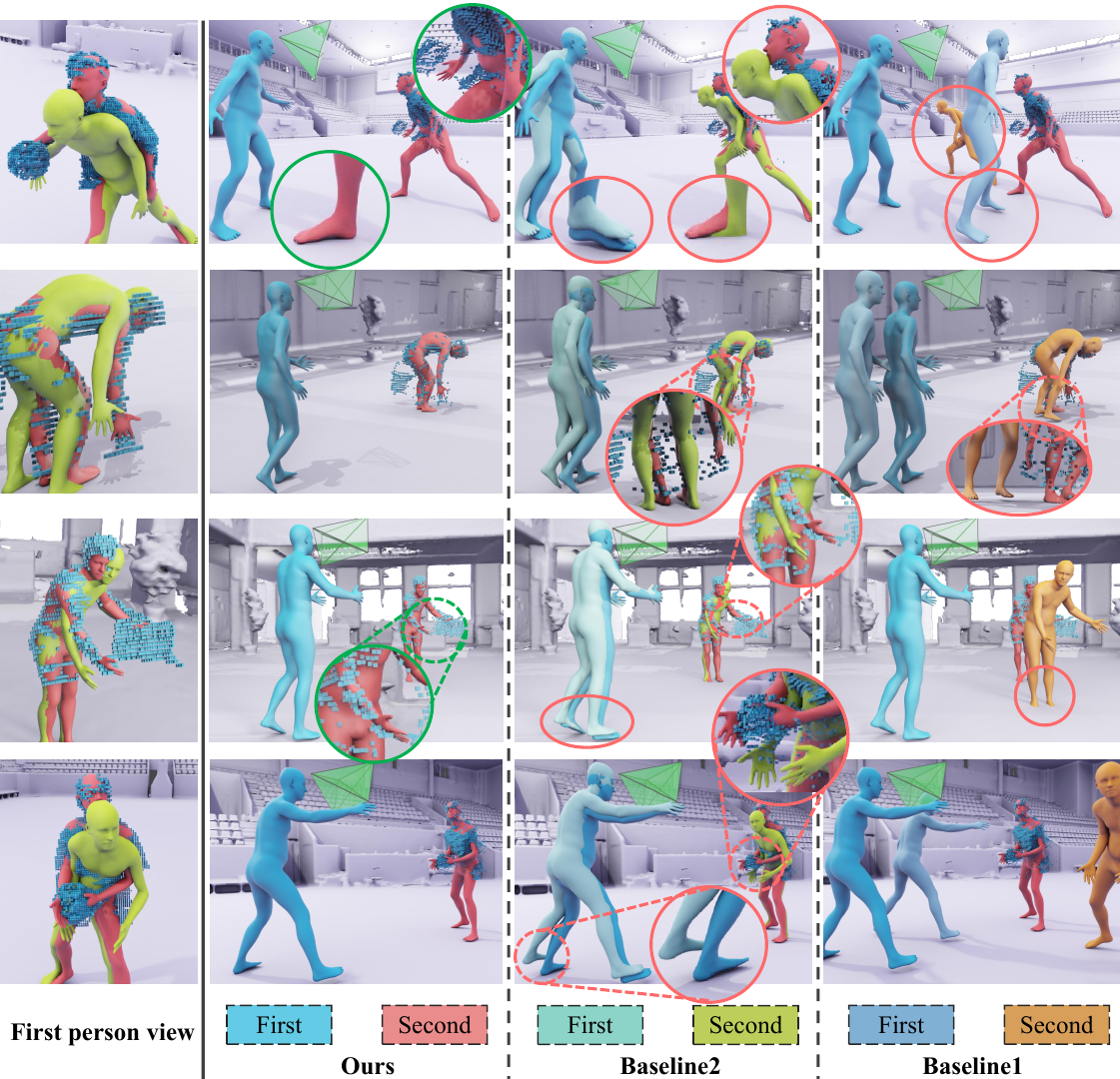}  
   \vspace{-2mm}
   \caption{Qualitative comparison between different methods. 
   The first column is the first-person view. The second column shows our results. The third column: Baseline2's results. The fourth column: Baseline1's results.}
   \vspace{-4mm}

   \label{fig:optimized_comparison}
\end{figure*}

In this section, we present our experiments and results. 
Firstly, we describe the implementation details of our capturing system and joint optimization in \cref{subsec:implementation}. 
Next, we introduce the evaluation metrics and present HiSC4D, which serves as a benchmark for LiDAR-based 3D human pose estimation from an egocentric view in \cref{subsec:metrics}. 
Finally, we compare our method with baselines (\cref{subsec:compare}) and evaluate the technical contributions of HiSC4D qualitatively and quantitatively (\cref{subsec:evaluation}) on various challenging scenarios. 
As depicted in \cref{fig:dataset}, our approach can capture challenging multi-human poses with interactions and even reconstruct large-scale scenes. 
Furthermore, \cref{fig:optimized_comparison} demonstrates the effectiveness of our human point cloud sequences in helping to address the occlusion challenges associated with the egocentric LiDAR data. 

\subsection{Implementation details}
\label{subsec:implementation}
Our experiments were executed on a system equipped with an NVIDIA GeForce GTX 3090Ti GPU and an Intel Core i7-10700K CPU to ensure optimal performance and accuracy. The optimization process, based on human-scene interaction, was conducted offline, post the preparation of all data.
We set the following empirically determined parameters: $\mathcal{L}_{cont}$ = 1, $\mathcal{L}_{sliding}$ = 1, $\mathcal{L}_{ort}$ = 1, $\mathcal{L}_{cont}$ = 2, $\mathcal{L}_{smt}$ = 4, $l = 200$, and we choose Adam \cite{adam15kingma} as the optimizer with learning rate set to 0.001. The optimization frame window, denoted as $k$, is set to 500. The iterations in optimization stages are set to 50, 40, and 110, respectively.

\subsection{Metrics}
\label{subsec:metrics}
To assess the plausibility of human-scene interactions, as well as pose smoothness, we utilize the following evaluation metrics: Foot Chamfer Distance (\bm{$CD_{foot}$}), Foot Sliding Error (\textbf{FSE}), Acceleration (\textbf{ACC}), and Global Localization Error \textbf{(GLE)}. Additionally, we evaluate the pose accuracy using the Chamfer Distance of SMPL Vertices to human points (\bm{$CD_{v2p}$}). 

{
\PAR{\bm{$CD_{foot}$}} is the stable foot's Chamfer distance to the ground surface, calculated in the same manner as $\mathcal{L}_{cont}$.}
{
\textbf{FSE} is the foot sliding distance between consecutive stable foot vertices, calculated in the same manner as $\mathcal{L}_{sld}$.}

{\PAR{ACC} is the pelvis acceleration. Its unit is meters per square second ($m/s_2$). This metric is used to assess the consistency and smoothness of human motion. The ACC is calculated as:
\begin{equation}
   ACC = \frac{1}{(n-2) * {\varDelta t}^2} \sum_{i=1+1}^{n-1} \| T_{i+1} - 2T_{i} + T_{i-1} \|_2^2, 
\end{equation}
where $n$ is the total frame number, $\varDelta t$ is the time between every two frames and $T_{i}$ is the $i$-th frame's pelvis translation.}

\PAR{GLE} is measured as the Euclidean distance between the human mesh model at the start frame and the human mesh model at the end frame. This metric is used to evaluate the accuracy of the first-person localization, where the LiDAR wearer starts and stops at the same location. This metric is only used on the first person.
The Euclidean distance is given by:
\begin{equation}
   GLE = \|F(M_1) - F(M_{n-1})\|_2^2,
    \label{equ:merit_gle}
\end{equation}
\noindent where $F({M}_{1})$ and $F({M}_{n-1})$ represent the first and the last human mesh model's foot position.

\PAR{\bm{$CD_{v2p}$}} is the 3D Chamfer Distance from visible human vertices to the corresponding human point cloud. 
To project the mesh, we first transform the mesh to the local LiDAR coordinate, and then we use Hidden Points Removal \cite{katz2007direct} to remove the invisible mesh vertices. Finally, we remove the vertices that are out of LiDAR's FOV and not compatible with LiDAR's resolution. This metric is only used on the second person that LiDAR scans.

\begin{equation}
   CD_{v2p} = \frac{1}{N} \sum_{i=1}^{N} \min_{j=1}^{M} \| \mathbf{p}_i - \mathbf{v}_j \|_2^2,
\end{equation}
\noindent where $N$ is the number of visible vertices in the human mesh, $\mathbf{p}_i$ is the $i$-th vertex in the mesh, $M$ is the number of LiDAR points, $\mathbf{v}_j$ is the $j$-th LiDAR point, and $\|\cdot\|$ denotes the Euclidean distance between two points.

\begin{table*}[tb!]
	\centering
    \renewcommand\arraystretch{1.2}
	\caption{\textbf{Quantitative Comparison between Baselines and Our Method:} We report the comparison across scene for all metrics including $CD_{foot}$, FSE, ACC, GLE, and $CD_{v2p}$. Lower values are better. The best outcomes are \textbf{bolded}. Results for the two individuals are presented in separate tables.}
	\centering
     
     \vspace{-3mm}
     \textbf{(a) The first-person quantitative comparison.}
     \vspace{1mm}

	\label{tab:local_error}
     \begin{tabular}{rr|rrrrrrrr|r}
          \toprule
          \textbf{Metric} & \textbf{Method} & \textbf{S1} & \textbf{S2} & \textbf{S3} & \textbf{S4} & \textbf{S5} & \textbf{S6} & \textbf{S7} & \textbf{S8} & \textbf{Avg.} \\
          \midrule
          \multirow{3}[2]{*}{$CD_{foot}$ ($mm$)} & Baseline1 & 42.9 & 29.8 & 46.3 & 25.8 & 27.1 & 22.1 & \textbf{5.4} & 43.1 & 28.8 \\
               & Baseline2 & 36.9 & 68.6 & 51.1 & 49.1 & 47.1 & 29.1 & 51.2 & 34.5 & 47.1 \\
               & Ours & \textbf{7.1} & \textbf{29.1} & \textbf{7.9} & \textbf{8.6} & \textbf{10.4} & \textbf{3.8} & 12.1 & \textbf{5.3} & \textbf{10.9} \\
          \midrule
          \multirow{3}[2]{*}{FSE ($mm$)} & Baseline1 & \textbf{8.0} & \textbf{5.5} & \textbf{4.5} & 17.4 & 8.2  & 12.3 & \textbf{6.8} & 18.4 & 10.9 \\
               & Baseline2 & 14.2 & 10.7 & 9.8  & 22.9 & 11.3 & 15.5 & 14.3 & 27.6 & 16.8 \\
               & Ours & 8.6  & 6.7  & 5.6  & \textbf{10.0} & \textbf{5.7} & \textbf{7.2} & 7.3  & \textbf{16.7} & \textbf{8.8} \\
          \midrule
          \multirow{3}[1]{*}{ACC ($m$/$s^2$)} & Baseline1 & 2.2  & 1.7  & 1.4  & 3.2  & 1.6  & 2.3  & 2.3  & 1.3  & 2.1 \\
               & Baseline2 & 2.6  & 1.9  & 1.9  & 3.8  & 1.9  & 2.7  & 2.4  & 1.5  & 2.5 \\
               & Ours & \textbf{2.0} & \textbf{1.5} & \textbf{1.3} & \textbf{2.6} & \textbf{1.2} & \textbf{1.8} & \textbf{1.9} & \textbf{0.9} & \textbf{1.8} \\
          \midrule
          \multirow{3}[1]{*}{GLE ($mm$)} & Baseline1 & 2787.6 & 730.2 & 666.9 & 1318.2 & 1550.5 & 1439.9 & 4329.1 & 7584.7 & 2602.1 \\
               & Baseline2 & 73.3 & \textbf{32.8} & 39.1 & 28.4 & 46.4 & \textbf{41.2} & 87.8 & 52.0 & 49.6 \\
               & Ours & \textbf{33.6} & 38.7 & \textbf{30.3} & \textbf{38.5} & \textbf{23.0} & 42.3 & \textbf{42.3} & \textbf{75.5} & \textbf{41.1} \\
          \bottomrule
     \end{tabular}%

     \vspace{1em}

     \textbf{(b) The second-person quantitative comparison.}
     \vspace{1mm}
     
	\label{tab:local_error_b}
     \begin{tabular}{rr|rrrrrrrr|r}
          \toprule
          \textbf{Metric} & \textbf{Method} & \textbf{S1} & \textbf{S2} & \textbf{S3} & \textbf{S4} & \textbf{S5} & \textbf{S6} & \textbf{S7} & \textbf{S8} & \textbf{Avg.} \\
          \midrule
          \multirow{3}[2]{*}{$CD_{foot}$ ($mm$)} & Baseline1 & 6.0  & 41.3 & 18.4 & 49.7 & 7.9  & 16.5 & 19.9 & 5.0  & 23.3 \\
               & Baseline2 & 75.7 & 59.7 & 11.9 & 54.1 & 37.5 & 60.6 & 68.3 & 31.1 & 51.0 \\
               & Ours & \textbf{5.2} & \textbf{5.4} & \textbf{2.9} & \textbf{5.8} & \textbf{7.4} & \textbf{3.4} & \textbf{4.3} & \textbf{5.0} & \textbf{5.0} \\
          \midrule
          \multirow{3}[2]{*}{FSE ($mm$)} & Baseline1 & \textbf{7.3} & 6.9  & \textbf{4.0} & 25.3 & 25.8 & 20.1 & 7.4  & 40.9 & 17.9 \\
               & Baseline2 & 29.2 & 29.8 & 8.9  & 35.5 & 38.3 & 29.3 & 33.2 & 50.9 & 33.0 \\
               & Ours & 7.8  & \textbf{5.9} & 4.5  & \textbf{11.5} & \textbf{13.1} & \textbf{8.4} & \textbf{6.6} & \textbf{21.7} & \textbf{10.2} \\
          \midrule
          \multirow{3}[2]{*}{ACC ($m$/$s^2$)} & Baseline1 & \textbf{2.3} & \textbf{2.2} & \textbf{1.3} & 4.9  & 5.6  & 4.3  & \textbf{2.1} & 2.5  & 3.2 \\
               & Baseline2 & 7.0  & 8.1  & 1.5  & 7.2  & 8.8  & 6.5  & 9.2  & 2.3  & 6.5 \\
               & Ours & 2.6  & 2.3  & 1.5  & \textbf{3.9} & \textbf{5.1} & \textbf{3.4} & 2.2  & \textbf{1.3} & \textbf{2.8} \\
          \midrule
          \multirow{3}[2]{*}{ $CD_{v2p}$ ($mm$)} & Baseline1 & 459.1 & 642.9 & 617.7 & 675.9 & 627.4 & 528.3 & 507.0 & 454.7 & 569.7 \\
               & Baseline2 & 61.6 & 58.1 & 40.7 & 68.1 & 61.7 & 80.1 & 68.6 & 76.8 & 65.1 \\
               & Ours & \textbf{29.9} & \textbf{30.3} & \textbf{33.1} & \textbf{28.2} & \textbf{28.9} & \textbf{37.8} & \textbf{33.2} & \textbf{30.1} & \textbf{31.0} \\
          \bottomrule
     \end{tabular}%
\end{table*}

\subsection{Comparison}
\label{subsec:compare}
To thoroughly assess the effectiveness of our approach, we compare it against other established baseline methods, conducting both qualitative and subsequent quantitative assessments to ensure a comprehensive evaluation. We aim to highlight the relative merits and the progressive contributions of our methodologies.

\PAR{Baseline1.}
We designate the method that relies solely on IMU data as Baseline1. 

\PAR{Baseline2.}
We term the method that cooperates IMU pose data with LiDAR localization as Baseline2. 
For the first-person motion, the localization is derived from the LiDAR SLAM results, while for the second-person motion, the localization is derived from the ICP results between the IMU pose and the cropped human point clouds. 
It also enables the evaluation of the benefits derived from combining these modalities.

\PAR{Qualitative Comparison.}
The qualitative comparison between our approach and Baseline1 and Baseline2 is shown in \cref{fig:optimized_comparison}. 
Baseline1 exhibits significant localization errors and inaccurate height estimations due to severe IMU drifting over time and the assumption of a horizontal plane for the tracked human. 
On the other hand, Baseline2 suffers from sliding errors caused by the non-rigid attachment of the LiDAR to the root joint of the human body and the mismatch between the body model and the actual person. 
In contrast, our approach leverages a robust joint optimization scheme that incorporates human-scene interaction, resulting in significantly improved robustness and smoothness, especially in challenging large-scale scenes.

Furthermore, our methodology surpasses HSC4D~\cite{Dai_2022_CVPR} by accurately capturing the 3D human motions of both interacting individuals. 
Moreover, the utilization of human point clouds as strong physical constraints enables us to provide precise constraints on local poses, which distinguishes our approach from HSC4D.
As shown in \cref{fig:optimized_comparison}, our approach achieves more accurate multiple human motions and produces physical-plausible results, benefiting from the utilization of temporal consistency constraints, human-scene interaction cues and physical constraints.
Importantly, our approach faithfully reconstructs both interacting humans and scenes from a first-person egocentric view, addressing a problem that has not been adequately addressed in existing approaches.

\PAR{Quantitative comparison.}
For quantitative comparison, we conduct a thorough analysis of the optimized human poses from both the first and second persons, comparing them with their respective baselines. 
To assess the outcomes for the first person involved in the interaction, we employed metrics including $CD_{foot}$, ACC, and GLE. Conversely, the evaluation of the results for the second person was conducted using metrics like $CD_{foot}$, ACC, and $CD_{v2p}$. We report the results in \cref{tab:local_error}.

Comparing the results in terms of Global Localization Error (GLE), it is evident that the method relying solely on IMU data (Baseline1) exhibits significant drifting. However, in Baseline2 by incorporating LiDAR-based localization with the initial pose provided by the IMU, the average GLE is reduced by 98.1\%. Furthermore, after applying our multi-stage optimization method, the average GLE is further reduced 17.1\%. Overall, these results demonstrate the effectiveness of our approach in lowering drifting and improving the accuracy of global localization.

To compare the scene plausibility using metrics such as $CD_{foot}$ and $CD_{v2p}$, we observe that the IMU-only approach (Baseline1) results in a significant misalignment between the SMPL mesh and the actual human point cloud, with a $CD_{v2p}$ of 569.7 mm. By incorporating IMU pose and performing ICP using the LiDAR-captured human point clouds (Baseline2), the $CD_{v2p}$ is reduced to 65.1 mm. Furthermore, after applying our optimization method, the $CD_{v2p}$ is further reduced by more than 47.7\%, indicating a substantial improvement in scene plausibility.

While Baseline2 improves both global localization accuracy and the alignment between poses and point clouds, the data in \cref{tab:local_error} indicates an increase in the overall average ACC and FSE. This suggests that integrating LiDAR might inadvertently reduce pose smoothness and coherence of local poses. 
In contrast, our method exhibits a substantial improvement: 62.2\% for ACC and 19.3\% for FSE on the first person, and 78.1\% for ACC and 43.0\% for FSE on the second person.

In conclusion, the comparative analysis underscores the superiority of our method. 
It not only augments pose smoothness, but also greatly improves global pose precision.

\subsection{Evaluation}
\label{subsec:evaluation}

We evaluate the technical contributions of our HiSC4D by ablating different loss terms. We classify these loss terms into four types: $\mathcal{L}_{smt}$, $\mathcal{L}_{self}$, $\mathcal{L}_{scene}$, and $\mathcal{L}_{phy}$. The $\mathcal{L}_{smt}$ includes all the smoothness terms, and $\mathcal{L}_{self}$ represents self-constraint loss, including foot sliding, self-penetration, and IMU pose prior. $\mathcal{L}_{scene}$ consists of scene-related losses, including foot contact loss $\mathcal{L}_{cont}$ and body collision loss $\mathcal{L}_{coll}$. Finally, $\mathcal{L}_{phy}$ includes losses derived from point cloud, such as $\mathcal{L}_{v2p}$ and $\mathcal{L}_{p2m}$ losses, as well as the LiDAR to head constraint $\mathcal{L}_{l2h}$. Before conducting the optimization process, the global localization of the poses is initialized by the LiDAR trajectory (first-person) and by performing ICP with corresponding human point clouds (second-person).

\begin{figure}[!tb]
    \centering
     \includegraphics[width=0.99\linewidth]{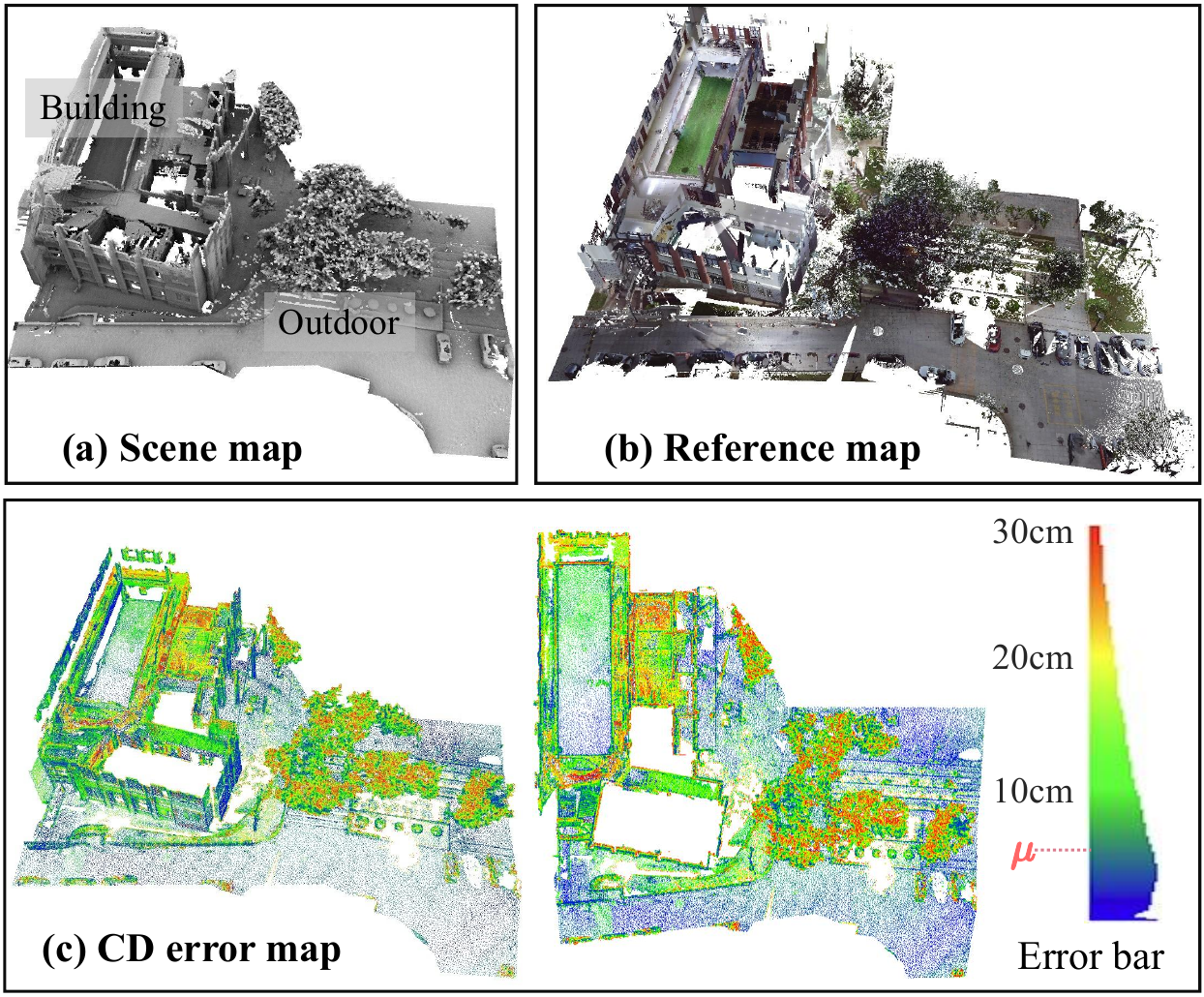}
     
     \vspace{-2mm}
     \caption{\textbf{Evaluation of our mapping result.} (a) Scene map from our method, (b) Reference colorful map, and (c) CD error (color-coded) map. The width of the color bar represents the points' quantity with this error. The mean CD error ($\mu$) is 6.5 cm.}
     \vspace{-3mm}

     \label{fig:mapping_error}
\end{figure}

\PAR{Scene mapping evaluation.} 
In our evaluation of sequence S3 as shown in \cref{fig:mapping_error}, we present the mapping results in a mixed indoor and outdoor environment. The sequence involves two individuals walking from inside a building, exploring the outdoor surroundings, and returning indoors.
The mean CD error is 6.5 cm and only a few points exceed 30 cm, so we truncated the points with errors larger than 30 cm on the figure to better visualize it.
We observed that the mapping results achieved high accuracy with minimal errors on the ground, typically below 5 cm (represented by blue points). However, larger errors were noticeable in regions containing outdoor trees and areas where the high-floored buildings were not traversed, as indicated by the CD error map.
In conclusion, our method demonstrates stable mapping performance in both indoor and outdoor environments, showcasing its effectiveness for a wide range of scenarios.

\begin{figure}[!tb]
    \centering
    \includegraphics[width=0.99\linewidth]{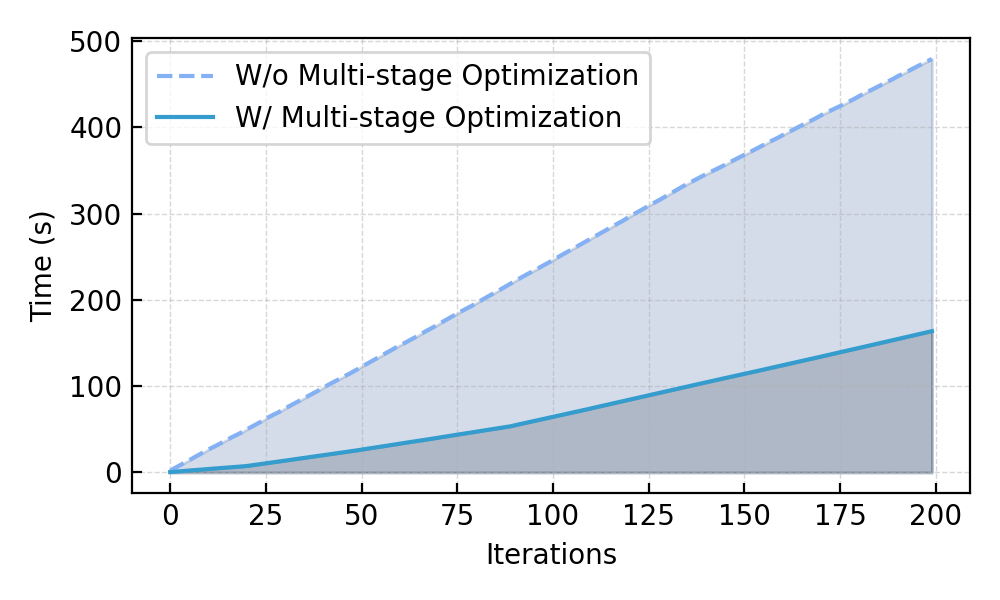}
    \vspace{-4mm}
    \includegraphics[width=0.99\linewidth]{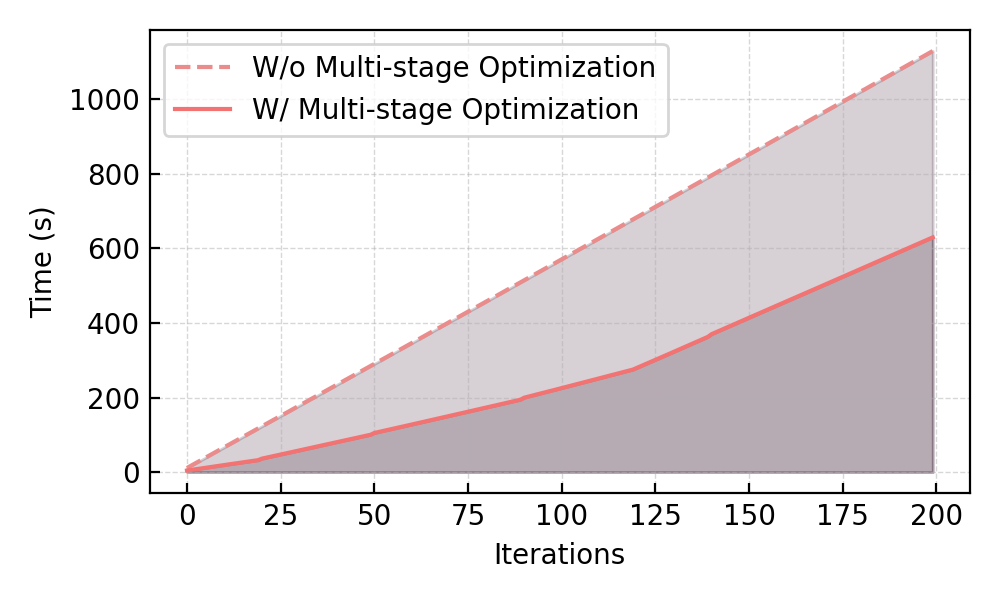}
    \vspace{-3mm}
    \caption{{Average runtime comparison between with and without multi-stage optimization over every 500 frames in sequence S4}
    }
    \vspace{-3mm}

    \label{fig:time}
\end{figure}

\begin{table}[tb!]
	\centering
    \renewcommand\arraystretch{1.2}
	\caption{{Runtime of our optimization method in S4 (7725 Frames) with and without multi-stage strategy.} "Iter" indicates the iteration number.}
	\centering
     
     \vspace{-2mm}
\begin{tabular}{ccccccc}
    \toprule
    \multirow{2}[2]{*}{With Multi-stage}    & \multicolumn{3}{c}{First person} & \multicolumn{3}{c}{Second person} \\
    \cmidrule(lr){2-4} \cmidrule(lr){5-7}   & Time & Iter. & Loss & Time & Iter. & Loss \\
    \midrule
    No          & 2.13 h & 200 & 53.79  &  5.02 h & 200 & 64.52 \\
    No          & \textbf{0.73 h} & 67  & 58.06 & \textbf{2.80 h} & 110 & 83.32 \\
    Yes         & 0.94 h & 250 & 54.13  & 3.76 h & 250 & \textbf{58.41} \\
    Yes (Ours)  & \textbf{0.73 h} & 200 & \textbf{53.56} & \textbf{2.80 h} & 200 & 68.71 \\
    \bottomrule
    \end{tabular}%
    \vspace{-3mm}
    
    \label{tab:time_eval}%
\end{table}

\begin{table*}[!htb]
    
    \caption{\textbf{Quantitative evaluation of our optimization method:} 
    Columns show the combination of loss terms used. We evaluate with metrics including $CD_{foot}$, ACC, GLE, $\mathcal{L}_{coll}$, and $CD_{v2p}$. Lower values are better. The best and second-best results are highlighted and underlined, respectively.}
    
    \centering

    \vspace{-3mm}
    \textbf{(a) The first-person quantitative evaluation.}
    \vspace{1mm}

\begin{tabular}{c|cccc|rrrrrrrr|r}
    \toprule
    \multicolumn{1}{r|}{\textbf{Metric}} & \multicolumn{1}{r}{\boldmath{}\textbf{$\mathcal{L}_{smt}$}\unboldmath{}} & \multicolumn{1}{r}{\boldmath{}\textbf{$\mathcal{L}_{self}$}\unboldmath{}} & \multicolumn{1}{r}{\boldmath{}\textbf{$\mathcal{L}_{scene}$}\unboldmath{}} & \multicolumn{1}{r|}{\boldmath{}\textbf{$\mathcal{L}_{phy}$}\unboldmath{}} & \textit{\textbf{S1}} & \textit{\textbf{S2}} & \textit{\textbf{S3}} & \textit{\textbf{S4}} & \textit{\textbf{S5}} & \textit{\textbf{S6}} & \textit{\textbf{S7}} & \textit{\textbf{S8}} & \textit{\textbf{Avg}} \\
    \midrule
    \multicolumn{1}{c|}{\multirow{5}[2]{*}{$CD_{foot}$\newline{} ($mm$)}} & \greencheck & \redx & \redx & \redx & 28.4 & 68.0 & 43.7 & 37.0 & 38.5 & 29.8 & 47.4 & 12.1 & 38.7 \\
        & \greencheck & \greencheck & \redx & \redx & 36.6 & 45.2 & 34.8 & 58.8 & 36.9 & 57.9 & 39.0 & 24.4 & 43.1 \\
        & \greencheck & \greencheck & \redx & \greencheck & 19.0 & 52.6 & 31.5 & 29.1 & 23.4 & 15.5 & 33.5 & 15.2 & 26.6 \\
        & \greencheck & \greencheck & \greencheck & \redx & \textbf{3.4} & \textbf{3.7} & \textbf{2.4} & \textbf{4.5} & \textbf{2.7} & \underline{4.0} & \textbf{3.3} & \textbf{4.0} & \textbf{3.6} \\
        & \greencheck & \greencheck & \greencheck & \greencheck & \underline{7.1} & \underline{29.1} & \underline{7.9} & \underline{8.6} & \underline{10.4} & \textbf{3.8} & \underline{12.1} & \underline{5.3} & \underline{10.9} \\
    \midrule
    \multicolumn{1}{c|}{\multirow{5}[2]{*}{ACC\newline{} ($m$/$s^2$)}} & \greencheck & \redx & \redx & \redx & \textbf{0.7} & \textbf{0.7} & \textbf{0.8} & \textbf{1.1} & \textbf{0.8} & \textbf{1.0} & \textbf{0.7} & \textbf{0.2} & \textbf{0.8} \\
        & \greencheck & \greencheck & \redx & \redx & 2.4 & 1.7 & 1.8 & 2.9 & 1.5 & 2.0 & 2.2 & 1.2 & 2.1 \\
        & \greencheck & \greencheck & \redx & \greencheck & \underline{1.8} & \underline{1.4} & 1.4 & \underline{2.5} & \underline{1.2} & \underline{1.7} & \underline{1.7} & \underline{0.9} & \underline{1.7} \\
        & \greencheck & \greencheck & \greencheck & \redx & 2.5 & 1.8 & 1.8 & 3.0 & 1.6 & 2.2 & 2.1 & 1.0 & 2.1 \\
        & \greencheck & \greencheck & \greencheck & \greencheck & 2.0 & 1.5 & \underline{1.3} & 2.6 & \underline{1.2} & 1.8 & 1.9 & \underline{0.9} & 1.8 \\
    \midrule
    \multicolumn{1}{c|}{\multirow{5}[2]{*}{$\mathcal{L}_{coll}$\newline{} ($mm$)}} & \greencheck & \redx & \redx & \redx & 0.8 & 0.3 & 4.6 & 0.7 & 1.3 & 3.7 & 0.2 & 0.1 & 1.2 \\
        & \greencheck & \greencheck & \redx & \redx & 0.6 & 8.6 & 1.6 & 19.7 & 6.3 & 16.0 & 5.1 & 3.7 & 8.8 \\
        & \greencheck & \greencheck & \redx & \greencheck & 0.7 & 0.2 & 8.4 & 0.8 & 1.0 & 0.4 & 0.1 & 0.1 & 1.2 \\
        & \greencheck & \greencheck & \greencheck & \redx & \textbf{0.0} & \textbf{0.0} & \textbf{0.0} & \textbf{0.0} & \textbf{0.0} & \textbf{0.0} & \textbf{0.0} & \textbf{0.0} & \textbf{0.0} \\
        & \greencheck & \greencheck & \greencheck & \greencheck & \textbf{0.0} & \textbf{0.0} & \textbf{0.0} & \textbf{0.0} & \textbf{0.0} & \textbf{0.0} & \textbf{0.0} & \textbf{0.0} & \textbf{0.0} \\
    \midrule
    \multicolumn{1}{c|}{\multirow{5}[2]{*}{GLE\newline{} ($mm$)}} & \greencheck & \redx & \redx & \redx & 59.0 & 44.7 & 51.6 & 59.4 & 37.0 & 56.1 & 91.2 & 54.8 & 58.6 \\
        & \greencheck & \greencheck & \redx & \redx & 99.6 & 153.0 & 90.7 & 387.8 & 105.9 & 163.4 & 103.8 & 236.4 & 192.4 \\
        & \greencheck & \greencheck & \redx & \greencheck & \underline{34.0} & \textbf{33.4} & \textbf{21.1} & \textbf{24.5} & \underline{25.7} & \textbf{41.0} & \textbf{42.3} & \textbf{63.1} & \textbf{35.2} \\
        & \greencheck & \greencheck & \greencheck & \redx & 128.4 & 155.6 & 206.7 & 402.6 & 151.9 & 297.0 & 129.1 & 300.4 & 236.7 \\
        & \greencheck & \greencheck & \greencheck & \greencheck & \textbf{33.6} & \underline{38.7} & \underline{30.3} & \underline{38.5} & \textbf{23.0} & \underline{42.3} & \textbf{42.3} & \underline{75.5} & \underline{41.1} \\
    \bottomrule
    \end{tabular}%
    
     \vspace{1em}
     \textbf{(b) The second-person quantitative evaluation.}
     \vspace{1mm}

\begin{tabular}{c|cccc|rrrrrrrr|r}
    \toprule
    \multicolumn{1}{r|}{\textbf{Metric}} & \multicolumn{1}{r}{\boldmath{}\textbf{$\mathcal{L}_{smt}$}\unboldmath{}} & \multicolumn{1}{r}{\boldmath{}\textbf{$\mathcal{L}_{self}$}\unboldmath{}} & \multicolumn{1}{r}{\boldmath{}\textbf{$\mathcal{L}_{scene}$}\unboldmath{}} & \multicolumn{1}{r|}{\boldmath{}\textbf{$\mathcal{L}_{phy}$}\unboldmath{}} & \textit{\textbf{S1}} & \textit{\textbf{S2}} & \textit{\textbf{S3}} & \textit{\textbf{S4}} & \textit{\textbf{S5}} & \textit{\textbf{S6}} & \textit{\textbf{S7}} & \textit{\textbf{S8}} & \textit{\textbf{Avg}} \\
    \midrule
    \multicolumn{1}{c|}{\multirow{5}[2]{*}{$CD_{foot}$\newline{} ($mm$)}} & \greencheck & \redx & \redx & \redx & 73.6 & 47.6 & 12.7 & 53.0 & 37.1 & 58.1 & 61.9 & 26.6 & 47.2 \\
        & \greencheck & \greencheck & \redx & \redx & 96.5 & 37.3 & 15.7 & 187.0 & 54.0 & 41.3 & 106.2 & 33.6 & 83.3 \\
        & \greencheck & \greencheck & \redx & \greencheck & 46.8 & 47.6 & 37.9 & 29.3 & 18.8 & 46.3 & 20.0 & 20.0 & 31.6 \\
        & \greencheck & \greencheck & \greencheck & \redx & \textbf{4.1} & \textbf{4.1} & \textbf{2.2} & \textbf{5.2} & \textbf{7.4} & \underline{6.0} & \textbf{4.1} & \underline{5.4} & \textbf{4.8} \\
        & \greencheck & \greencheck & \greencheck & \greencheck & \underline{5.2} & \underline{5.4} & \underline{2.9} & \underline{5.8} & \textbf{7.4} & \textbf{3.4} & \underline{4.3} & \textbf{5.0} & \underline{5.0} \\
    \midrule
    \multicolumn{1}{c|}{\multirow{5}[2]{*}{ACC\newline{} ($m$/$s^2$)}} & \greencheck & \redx & \redx & \redx & \textbf{0.9} & \textbf{0.8} & \textbf{0.7} & \textbf{1.3} & \textbf{1.6} & \textbf{1.4} & \textbf{0.8} & \textbf{0.3} & \textbf{1.0} \\
        & \greencheck & \greencheck & \redx & \redx & \underline{2.0} & \underline{1.7} & \underline{1.2} & \underline{3.7} & \underline{3.8} & \underline{2.8} & \underline{1.9} & \underline{1.5} & \underline{2.4} \\
        & \greencheck & \greencheck & \redx & \greencheck & 2.4 & 2.0 & 1.5 & 3.7 & 4.3 & 3.5 & 2.3 & 1.4 & 2.7 \\
        & \greencheck & \greencheck & \greencheck & \redx & 2.4 & 2.2 & 1.3 & 3.9 & 4.6 & 3.1 & 2.1 & 1.4 & 2.7 \\
        & \greencheck & \greencheck & \greencheck & \greencheck & 2.6 & 2.3 & 1.5 & 3.9 & 5.1 & 3.4 & 2.2 & 1.3 & 2.8 \\
    \midrule
    \multicolumn{1}{c|}{\multirow{5}[2]{*}{$\mathcal{L}_{coll}$\newline{} ($mm$)}} & \greencheck & \redx & \redx & \redx & 31.2 & 6.5 & 0.8 & 15.0 & 5.0 & 17.3 & 13.0 & 3.3 & 11.4 \\
        & \greencheck & \greencheck & \redx & \redx & 42.2 & 5.4 & 1.8 & 84.0 & 18.0 & 8.3 & 6.5 & 6.7 & 25.8 \\
        & \greencheck & \greencheck & \redx & \greencheck & 15.4 & 6.0 & 3.5 & 2.6 & 1.3 & 8.0 & 1.2 & 2.0 & 4.4 \\
        & \greencheck & \greencheck & \greencheck & \redx & \textbf{0.0} & \textbf{0.0} & \textbf{0.0} & \textbf{0.0} & \textbf{0.0} & \textbf{0.0} & \textbf{0.0} & \textbf{0.0} & \textbf{0.0} \\
        & \greencheck & \greencheck & \greencheck & \greencheck & \textbf{0.0} & \textbf{0.0} & \textbf{0.0} & \textbf{0.0} & \textbf{0.0} & \textbf{0.0} & \textbf{0.0} & \textbf{0.0} & \textbf{0.0} \\
    \midrule
    \multicolumn{1}{c|}{\multirow{5}[2]{*}{$CD_{v2p}$\newline{} ($mm$)}} & \greencheck & \redx & \redx & \redx & 61.9 & 57.2 & 43.0 & 71.1 & 66.2 & 82.0 & 83.3 & 80.5 & 69.4 \\
        & \greencheck & \greencheck & \redx & \redx & 71.6 & 64.4 & 46.7 & 106.6 & 109.5 & 96.3 & 129.1 & 136.5 & 99.4 \\
        & \greencheck & \greencheck & \redx & \greencheck & \textbf{29.3} & \underline{45.0} & \underline{44.7} & \textbf{28.1} & \textbf{28.3} & \textbf{35.9} & \underline{52.1} & \textbf{30.0} & \underline{36.9} \\
        & \greencheck & \greencheck & \greencheck & \redx & 144.8 & 105.8 & 69.4 & 82.7 & 106.6 & 114.0 & 176.3 & 170.2 & 122.9 \\
        & \greencheck & \greencheck & \greencheck & \greencheck & \underline{29.9} & \textbf{30.3} & \textbf{33.1} & \underline{28.2} & \underline{28.9} & \underline{37.8} & \textbf{33.2} & \underline{30.1} & \textbf{31.0} \\
    \bottomrule
    \end{tabular}%
    \vspace{-3mm}
     
    \label{tab:quanti_eva}
    
\end{table*}

\PAR{Runtime evaluation.}
{As shown in \cref{fig:time}, the }multi-stage optimization strategy consistently demonstrates faster convergence and lower loss. {According to \cref{tab:time_eval}, our method requires 0.73 hours for the first person and 2.80 hours for the second person. 
Compared to the optimization without the multi-stage strategy, which takes 2.13 hours for the first person and 5.02 hours for the second person, our method reduces the runtime by 65.73\% for the first person and 44.22\% for the second perso.
Furthermore, as seen from} the second and the fourth row in \cref{tab:time_eval}, with the same optimization time, the losses with multi-stage strategy are 7.8\% and 17.5\% less than those without the multi-stage strategy for the two persons, respectively. 
{These results demonstrate that our sequential optimization strategy—starting with global translation, then integrating global orientation, and finally optimizing human body pose—not only preserves accuracy but also significantly accelerates the process.}

\PAR{Quantitative evaluation.} {To quantitatively evaluate our method, we assess the error in the proposed dataset by reporting} $CD_{foot}$, ACC, $\mathcal{L}_{coll}$, GLE and $CD_{v2p}$ with {different loss terms ablated. This approach demonstrates the effectiveness of each component of our method.}
As shown in \cref{tab:quanti_eva}, when ablating other loss terms with only using $\mathcal{L}_{smt}$, the acceleration (ACC) is the lowest among all sequences. Yet, its performance on scene-related metrics is worse. Metrics like GLE and $CD_{v2p}$ are adversely impacted. Moreover, a notably high $CD_{foot}$ value indicates that the poses are not scene-plausible.
From the fourth column in each block of the table, we observe that when incorporating $\mathcal{L}_{scene}$ but without $\mathcal{L}_{phy}$, the $CD_{foot}$ achieves optimal performance, indicating a consistent foot-ground contact. However, in contrast, the $CD_{v2p}$ and GLE values are the worst compared to other methods. This reveals a misalignment between the poses and the human point cloud and diminished global localization accuracy.
These findings highlight that merely depending on scene constraints like contacts or penetration is inadequate for generating realistic human motions. Without external cues, like point clouds or LiDAR-based localization which can offer precise human global positioning, the estimated human poses may diverge from the actual observed values in real-world scenarios, thereby affecting the pose quality.

In conclusion, In every scenario where a loss term is ablated, even if the method achieves the best in a particular metric, it inevitably exhibits significant shortcomings in another.
In contrast, our method integrates all the proposed loss terms,  not only excels but dominates in key metrics. Specifically, it stands optimal in metrics like $CD_{v2p}$, which measures the alignment between poses and corresponding human point clouds. 
Furthermore, our approach is a close runner-up in other pivotal metrics such as $CD_{foot}$ and GLE. 
This superior performance underscores the indispensable nature of our proposed constraint terms. 
The evaluations unequivocally attest to the prowess of our optimization method, which adeptly harnesses the synergy of all loss terms, ushering in a marked elevation in overall performance.

\begin{figure*}[!tbp]
    \centering
    \includegraphics[width=0.96\linewidth]{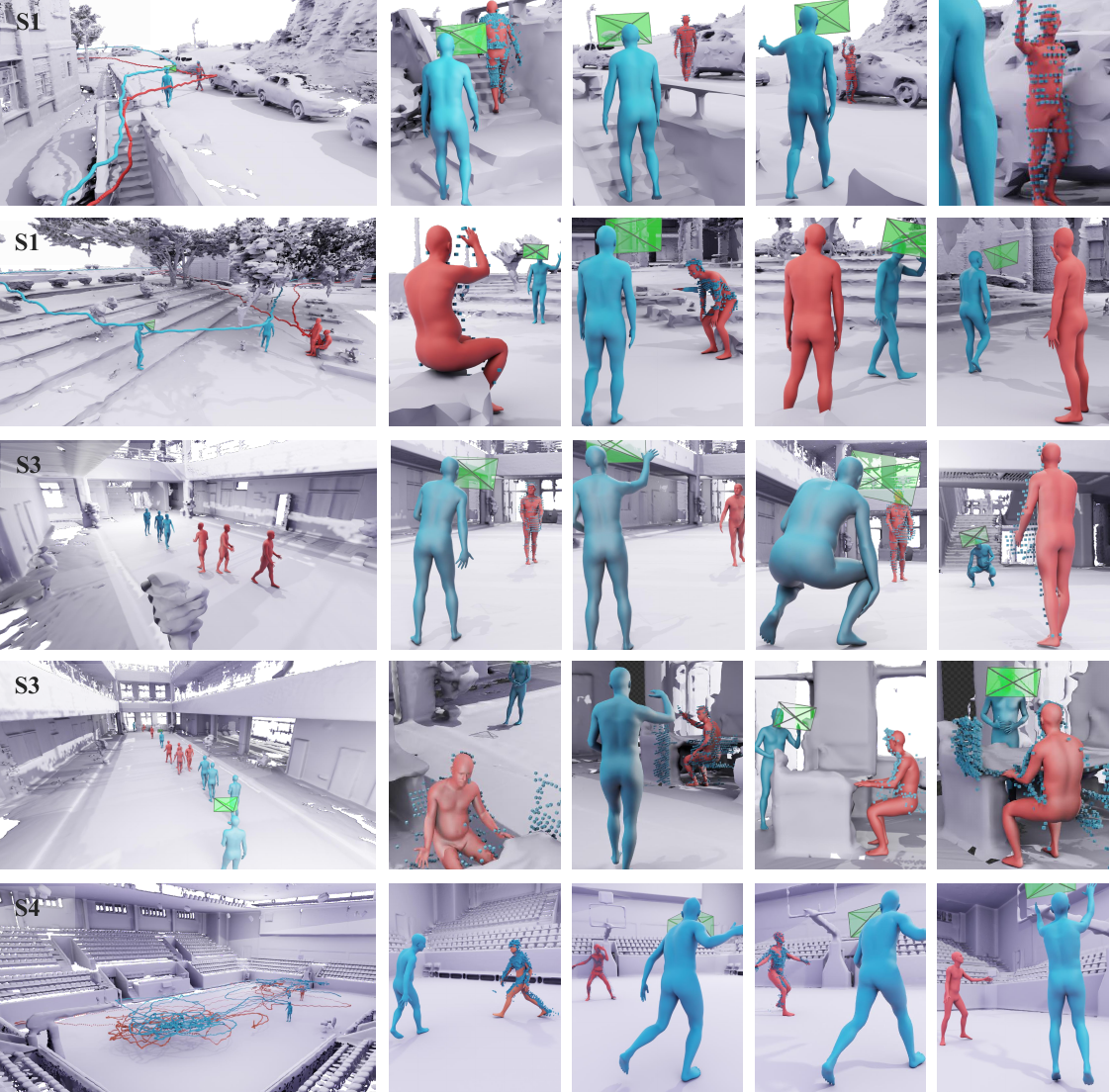}  
   \vspace{-2mm}
    \caption{\textbf{Results gallery.} Each row in the figure corresponds to a specific scene, with the first column indicating the overview of the scene. The second to fifth columns provide a close view of the human motions captured within each scene. 
    }
    \vspace{-3mm}
    \label{fig:dataset}
\end{figure*}

\PAR{Qualitative evaluation.}
\begin{figure}[!tbp]
    \centering
    \includegraphics[width=0.96\linewidth]{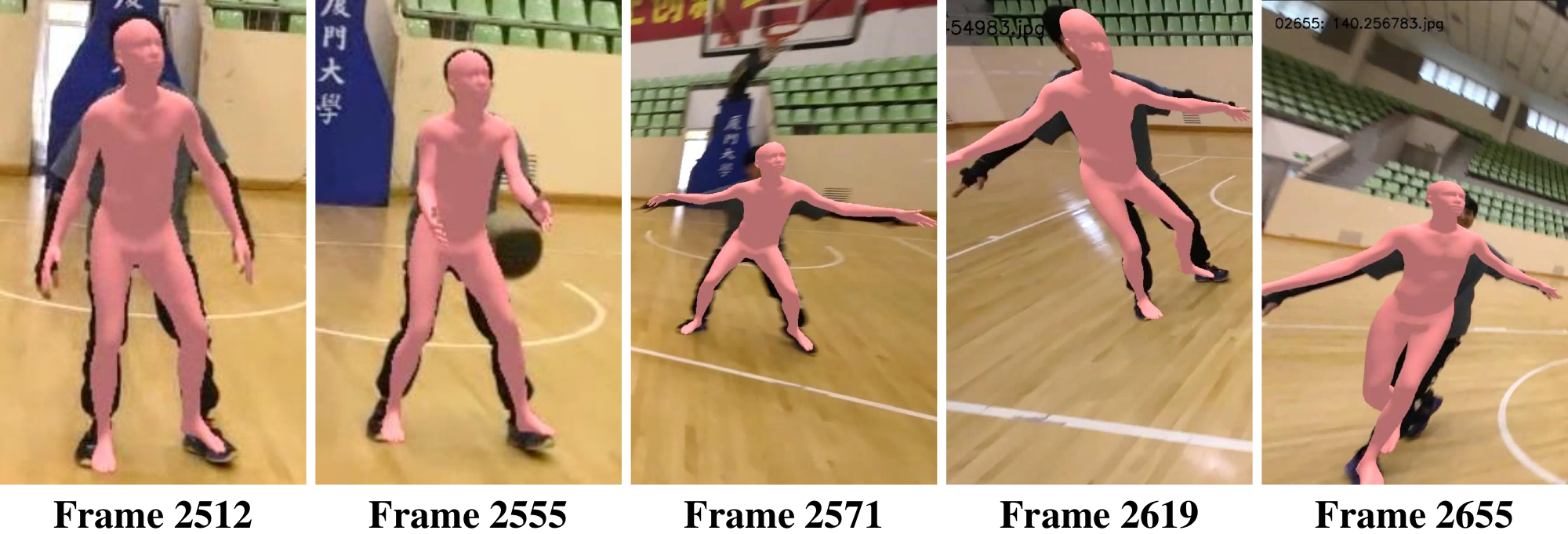}  
    \vspace{-2mm}
    \caption{
        Qualitative results of the second person's motions projected to cameras
    }
    \vspace{-3mm}

    \label{fig:quali_eva}
 \end{figure}
{For the qualitative evaluation of local poses, we utilized an action camera positioned near the LiDAR while capturing the sequence S4. The camera was manually calibrated and synchronized with the LiDAR, allowing us to project human motions onto the images. As shown in \cref{fig:quali_eva}, the human mesh aligns well with the image.}
In \cref{fig:dataset}, we provide qualitative examples illustrating the effectiveness of our method in various human-centric situations.
The first two rows depict two individuals transitioning from indoors to the outdoors, with actions such as greetings and sitting. These scenarios highlight our method's adeptness in accurately representing human behaviors across changing environments.
The third and fourth rows portray a spacious indoor setting. They present scenarios like participants in a throwing game and individuals exchanging greetings near an office desk. These examples show our method's strength in recognizing everyday human activities indoors.
The last row captures a basketball game in an indoor gymnasium, emphasizing moments of dribbling, face-offs, and shooting attempts. This vividly underlines our method's precision in recording rapid, interactive human motions.
In summary, these examples demonstrate the comprehensive ability and flexibility of our method, capable of capturing a wide range of human motions, from simple indoor tasks to dynamic outdoor sports.

\subsection{Limitations}
As a trial for human-centered 4D human and scene capture in large-scale scenes. We have demonstrated compelling reconstruction results. Nonetheless, our approach is subject to some limitations. 
One limitation of this study is the vertical field of view limitation of the LiDAR, which is 33.2° / 45°. This results in more severe body truncations and joint occlusions for the second person. Additionally, due to the low resolution of the LiDAR sensor, our method cannot reconstruct human textures.
To address these limitations, it is promising to incorporate additional modalities, such as RGB video, which can provide more detailed information for generating geometry details of the second person.
The second limitation is the computational complexity. While the scene-aware optimization algorithm in our proposed method offers significant benefits in terms of accuracy and realism, it also introduces computational complexity. This complexity can pose challenges in real-time and large-scale applications.
Finally, while HiSC4D leverages the built-in IMU of the LiDAR sensor to improve SLAM performance, there are limitations in extreme situations such as long-duration occlusions. 
Future research can try to incorporate the wearing IMUs to address these specific cases and enhance the overall robustness of the SLAM system.

\section{Conclusion}
\label{sec:Conclusion}
We presented HiSC4D, a Human-centered 4D Human and scene capture method to accurately and efficiently create a dynamic digital world using only body-mounted IMUs and a head-mounted LiDAR. 
Specifically, we propose a scene-aware optimization algorithm that integrates initial IMU-based human poses, human point clouds, and physical environmental cues from scenes, enabling accurate and smooth human motions during interaction in large-scale real-world environments.
Additionally, we introduce a dataset containing large scenes and challenging human motions, enriched with diverse social interactions and 3D annotations. Compared to HSC4D, by cooperating with the built-in IMU of the LiDAR sensor, HiSC4D exhibits a more robust simultaneous localization and mapping (SLAM) capability in various scenarios, including activities like playing basketball.
Extensive experimental results validate the effectiveness and robustness of our approach for 4D human and scene capturing, covering a wide range of challenging scenarios. These results demonstrate our method's capability to faithfully capture and reconstruct interacting human motions across diverse scenarios.
In conclusion, our methodology, backed by a robust optimization scheme and comprehensive evaluation metrics, presents a significant advancement in capturing human interactions in large scale spaces. The detailed insights into the implementation and the comparative analysis with baselines underscore the accuracy and reliability of our approach. We believe this work will foster the creation of a dynamic digital world with broad applications, such as social interaction analysis, long-term human motion analysis, human activity recognition, etc.

\appendices

\ifCLASSOPTIONcaptionsoff
  \newpage
\fi

\bibliographystyle{IEEEtran}
\bibliography{IEEEabrv,egbib}

\end{document}